\documentclass{article}

 \usepackage[preprint]{neurips_2026}

\usepackage[utf8]{inputenc} 
\usepackage[T1]{fontenc}    
\usepackage{hyperref}       
\usepackage{url}            
\usepackage{booktabs}       
\usepackage{amsfonts}       
\usepackage{nicefrac}       
\usepackage{microtype}      
\usepackage{xcolor}         
\usepackage{graphicx}
\usepackage{amsmath}
\usepackage{amssymb}
\usepackage{siunitx}
\newcommand{\std}[1]{$\pm$#1}
\usepackage{multirow}
\usepackage{wrapfig}
\usepackage{placeins}

\usepackage{caption}

\title{PhAME: Phenotype-Aware Molecular Editing\\via Latent Diffusion} 

\author{%
  \textbf{\L{}ukasz Janisi\'{o}w$^{1,2}$\thanks{Equal contribution.}} \qquad
  \textbf{Sebastian Musia\l{}$^{1 \ast}$} \qquad
  \textbf{Bartosz Zieli\'{n}ski$^{1,3}$} \\
  \vspace{0pt} \\ 
  \textbf{Dawid Rymarczyk$^{1,5    \dagger}$} \qquad
  \textbf{Tomasz Danel$^{4}$\thanks{Corresponding authors: \texttt{\{tomasz.danel, dawid.rymarczyk\}@uj.edu.pl}}} \\
  \vspace{0pt} \\
  \textsuperscript{1}Faculty of Mathematics and Computer Science, Jagiellonian University \\
  \textsuperscript{2}Doctoral School of Exact and Natural Sciences, Jagiellonian University \\
  \textsuperscript{3}Jagiellonian Center for Artificial Intelligence, Jagiellonian University \\
  \textsuperscript{4}Faculty of Chemistry, Jagiellonian University \quad \textsuperscript{5}Ardigen SA
}

\begin{document}

\maketitle

\begin{abstract}

Small-molecule drug discovery requires simultaneous optimization of numerous properties of candidate molecules. These properties can be investigated through the analysis of high-dimensional biological signatures, such as cell morphology and transcriptomic perturbations, which provide a rich perspective on the underlying biological mechanisms. However, existing generative methods, which use those signatures for optimization, fail to meet two key requirements: providing precise guidance toward desired phenotypic signatures while maintaining structural proximity to a known hit. We introduce PhAME (Phenotype-Aware Molecular Editing), a latent diffusion framework that overcomes this challenge by recasting molecular optimization as editing in the latent space of a pretrained graph-based VAE. Our central contribution is a compositional classifier-free guidance scheme with two independent scales, one for the phenotype-conditioning and one for similarity to the seed structure, allowing practitioners to control the tradeoff between these two objectives. Empirical evaluations across diverse benchmarks, including docking score optimization and multimodal phenotypic generation, demonstrate that PhAME achieves state-of-the-art results while maintaining high chemical validity and novelty.
\end{abstract}

\section{Introduction}

  

\begin{wrapfigure}{r}{0.5\textwidth}
\vspace{-15.0mm}
\includegraphics[width=\linewidth]{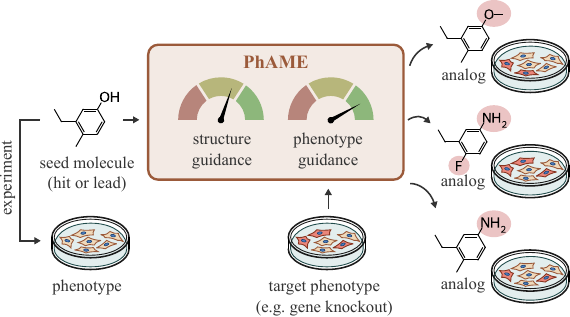}
\caption{\textbf{Molecular editing with PhAME.} Given a seed molecule and target phenotype, the model generates close structural analogs that induce a target phenotype by balancing two independent guidance scales: structure and phenotype.}
\label{fig:teaser}
\vspace{-8.0mm}
\end{wrapfigure}

Drug discovery is a search over a chemical  space estimated  at $10^{60}$ drug-like molecules~\citep{polishchuk2013estimation}, but in practice it rarely starts from scratch. The dominant bottleneck is the iterative refinement of an 
already-active compound through \textit{Hit-to-Lead} (H2L) and \textit{Lead Optimization} (LO), reshaping a screening hit into a potent, drug-like, synthesizable candidate~\citep{niazi2023coming}. Compounding this, the field is shifting from single-target binding assays toward 
\textit{phenotypic} readouts, such as gene-expression changes and cell-morphology 
profiles, that capture how a molecule perturbs a biological system as a whole~\citep{chandrasekaran2021image, swinney2013phenotypic}.
A generative model that can both edit a known hit and steer it toward such system-level signals would be a powerful tool for ML-driven discovery, yet no method today delivers both.

Ligand-centric generators~\citep{gomez2018automatic, jin2018junction, jing2022torsional} produce valid molecules but ignore cell-level response and offer no mechanism to anchor on a seed compound. Phenotype-conditioned generators close the first gap but not the second: GxVAEs~\citep{li2024gxvaes} and SmilesGEN~\citep{liu2025phenotypic} both generate \textit{de novo} from a phenotypic target and cannot be steered around a known hit. A concurrent work~\citep{song2025phenomoler} adds molecule editing via discrete mask-and-reconstruct on SELFIES substructures, but only under L1000 transcriptomics. Even classifier-free guidance (CFG)~\citep{ho2022classifierfreediffusionguidance}, the standard steering tool for diffusion models, does not solve the problem. Its single guidance scale forces a trade-off between matching the target phenotype and staying close to the seed molecule.

We propose \textbf{PhAME} (\textbf{Ph}enotype-\textbf{A}ware \textbf{M}olecular \textbf{E}diting) to close both gaps: conditioning on a desired phenotype and anchoring on a seed compound. PhAME reformulates design as an editing task in the latent space of a frozen graph-based VAE~\citep{Nguyen2025VAE}. A seed molecule is partially noised and denoised under guidance, so the noise level controls the structural exploration radius. Structural proximity is measured in a learned embedding space of InfoAlign~\citep{liu2025learning} that reflects cellular response rather than raw chemical similarity. Crucially, we replace standard CFG with a \textit{compositional} variant that splits the guidance signal into two terms with \emph{independent} scales, one for the phenotypic target and one for structural anchoring, directly removing the trade-off baked into single-scale CFG.

Across docking-score optimization, transcriptomic-conditioned generation, and phenotype-driven mode of action (MoA) design, PhAME delivers the strongest results to date: the best novel hit ratio on four out of five MOOD~\citep{lee2023exploring} protein targets, and consistent superiority on phenotype-driven MoA retrieval, all while maintaining near-perfect validity and novelty.

Our contributions can be summarized as follows:
\begin{itemize}
    \item \textbf{Latent-diffusion molecular editing:} We introduce PhAME, a framework utilizing latent diffusion on graph-based embeddings to enable systematic Hit-to-Lead guided by complex biological signatures.
    \item \textbf{Compositional guidance mechanism:} We generalize classifier-free guidance to use two independent scales, one for phenotype matching and one for seed proximity, replacing the forced trade-off with explicit control over each objective.
    \item \textbf{SOTA across modalities:} PhAME sets new state-of-the-art on docking-score optimization, transcriptomic-conditioned generation, and Cell-Painting-driven MoA generation, with near-perfect validity and novelty.
\end{itemize}

The source code and accompanying data are available at \url{https://github.com/gmum/PhAME}.

\section{Related Work}

\textbf{Generative models for molecules.}
Molecular generation has progressed from SMILES-based VAEs~\citep{gomez2018automatic} to graph-based VAEs~\citep{jin2018junction}, and more recently to latent diffusion models~\cite{rombach2022high} that compress denoising into a learned manifold. Mol-CycleGAN~\cite{maziarka2020mol} already showed that the latent space of a graph VAE supports direct molecular optimization. Equivariant and graph variants of latent diffusion in chemistry include EDM~\cite{hoogeboom2022equivariant} and GCLDM~\cite{zhang2025geometry}. For property-driven optimization, score-based MOOD~\cite{lee2023exploring}, the flow-matching frameworks PropMolFlow~\cite{zeng2026propmolflow} and MoltenFlow~\cite{lobo2026property}, and fragment-level discrete diffusion with GenMol~\citep{lee2025genmol} offer stable alternatives to reinforcement learning~\cite{loeffler2024reinvent}. Conditioning is typically realized via classifier-free guidance (CFG)~\citep{ho2022classifierfreediffusionguidance,weiss2023guided}. Recent factorized variants in scPPDM~\cite{liang2025scppdm} and the multi-attribute CFGen~\cite{palma2024multi} expose multiple interpretable controls, but were developed for single-cell rather than molecular targets. PhAME instantiates a latent-diffusion-on-graph-VAE recipe on top of the DGVAE~\citep{Nguyen2025VAE} backbone and replaces single-scale CFG with a compositional scheme that scales the phenotypic target and structural anchor independently.

\textbf{Phenotype-conditioned molecular generation.}
A complementary line conditions generation on system-level biological readouts. Most methods consume L1000 transcriptomic signatures: the GAN of M\'endez-Lucio et al.~\cite{mendez2020novo}, the VAEs Triomphe~\cite{kaitoh2021triomphe}, GxVAE~\citep{li2024gxvaes} and SmilesGEN~\citep{liu2025phenotypic}, the graph latent diffusion GLDM~\cite{wang2024gldm}, and the chemistry LLM GEMGen~\cite{jiang2026phenotype}. A smaller group conditions on Cell Painting morphology: CPMolGAN~\citep{zapata2023cell}, MGMG~\citep{tang2025mgmg}, and the GFlowNet of Lu et al.~\cite{lu2024cell}. However, most of these methods generate molecules \textit{de novo}. PhAME differs by editing in a graph-VAE latent, anchoring proximity through InfoAlign~\cite{liu2025learning} embeddings aligned with both Cell Painting and L1000 via an information bottleneck, and exposing independent phenotypic and structural CFG scales.

\section{Methods}

\subsection{Background}
\label{sec:background}

\paragraph{Latent diffusion models.} Diffusion models~\citep{ho2020DDPM,song2021score} learn to generate data by reversing a gradual noising process. Operating this process directly on high-dimensional or discrete data, such as SMILES strings or molecular graphs, is computationally expensive and requires careful handling of discrete structures. Latent diffusion models (LDMs)~\citep{rombach2022high} resolve this by first training an autoencoder to map data into a compact continuous latent space, and then applying the diffusion process entirely within that space. In our setting, a pretrained SMILES VAE with encoder $\mathcal{E}$ and decoder $\mathcal{D}$ provides the latent space: $z_0 = \mathcal{E}(m)$ for a molecule $m \in \mathcal{M}$.

The forward process progressively corrupts $z_0$ into Gaussian noise via $q(z_t \mid z_0) = \mathcal{N}(z_t;\, \sqrt{\bar{\alpha}_t}\, z_0,\, (1 - \bar{\alpha}_t)\, I)$, where $\bar{\alpha}_t = \prod_{s=1}^{t} \alpha_s$ follows a predefined variance schedule and $z_T \approx \mathcal{N}(0, I)$ for sufficiently large $T$. A neural network $\epsilon_\theta(z_t, t)$ is trained to reverse this process by predicting the added noise, minimizing
\begin{equation}
    \mathcal{L} = \mathbb{E}_{z_0, \epsilon, t}\!\left[\left\| \epsilon - \epsilon_\theta(z_t, t) \right\|^2\right], \qquad \epsilon \sim \mathcal{N}(0, I),\; t \sim \mathcal{U}\{1, \dots, T\}.
\end{equation}

At inference, new molecules are generated by sampling $\hat{z}_T \sim \mathcal{N}(0, I)$ and iteratively denoising:
\begin{equation}
 \hat{z}_{t-1} = \frac{1}{\sqrt{\alpha_t}} \left( \hat{z}_t - \frac{1 - \alpha_t}{\sqrt{1 - \bar{\alpha}_t}}\, \epsilon_\theta(\hat{z}_t, t) \right) + \sigma_t\, \eta, \qquad \eta \sim \mathcal{N}(0, I),   
\end{equation}
where $\sigma_t$ controls sampling stochasticity ($\sigma_t = 0$ recovers DDIM deterministic sampling~\citep{song2021denoising}). The final molecule is obtained by decoding: $m'=\mathcal{D}(\hat{z}_0)$.

\paragraph{Classifier-free guidance.} To enable conditional generation without a separate classifier, classifier-free guidance (CFG)~\citep{ho2022classifierfreediffusionguidance} trains a single network $\epsilon_\theta(\hat{z}_t, t, c)$ on both conditional and unconditional objectives. During training, the conditioning signal $c$ is randomly replaced with a null token $\varnothing$ with probability $p_{\text{uncond}}$, so that the network implicitly learns both the conditional and unconditional score functions. At inference, the noise prediction is interpolated as
\begin{equation}
    \tilde{\epsilon}_\theta(\hat{z}_t, t, c) = \epsilon_\theta(\hat{z}_t, t, \varnothing) + w \left[\epsilon_\theta(\hat{z}_t, t, c) - \epsilon_\theta(\hat{z}_t, t, \varnothing)\right],
\end{equation}
where $w \geq 0$ is the guidance scale. Setting $w = 0$ recovers unconditional sampling; $w = 1$ corresponds to standard conditional sampling; and $w > 1$ amplifies the influence of the conditioning signal at the expense of sample diversity. This mechanism is agnostic to the dimensionality and semantics of $c$, making it naturally suited for conditioning on signals ranging from scalar properties to high-dimensional phenotypic profiles.

\subsection{Problem Definition}
\label{sec:problem}

Let $\mathcal{M}$ denote the space of valid molecules represented as SMILES strings. We write $m \in \mathcal{M}$ for a seed (lead) molecule and define a \textit{target signature} $c \in \mathbb{R}^K$, where $K \geq 1$. Depending on the task, $c$ may represent a scalar property target, a multi-property vector, a transcriptomic profile, or a cell morphology signature. Let $\phi : \mathcal{M} \to \mathbb{R}^K$ denote a (generally unknown) oracle mapping each molecule to its ground-truth signature, and let $\mathrm{sim} : \mathcal{M} \times \mathcal{M} \to [0, 1]$ be a structural similarity measure (e.g.\ Tanimoto similarity over Morgan fingerprints).

\paragraph{Objective.} Given a seed molecule $m$ and a target signature $c$, we seek an optimized molecule $m^{*}$ that minimizes divergence from $c$ while remaining structurally related to the seed:

\begin{equation}
    m^{*} = \arg\min_{m' \in \mathcal{M}} \; d\!\left(\phi(m'),\, c\right) + \lambda (1 - \mathrm{sim}(m, m')),
\end{equation}

where $m' \in \mathcal{M}$ denotes a candidate molecule, $d$ is a divergence in the signature space, and $\lambda \geq 0$ governs the trade-off between property improvement and structural conservation. Setting $\lambda = 0$ recovers unconstrained de novo generation.

In practice, we do not optimize this objective directly. The trade-off is controlled implicitly through the \textit{noise level} $t$ in a partial diffusion process, which governs the structural exploration radius around the seed, and the \textit{guidance scale} $w$ in classifier-free guidance (Section~\ref{sec:background}), which governs the strength of conditioning toward $c$.

\begin{figure}
    \centering
    \includegraphics[width=\linewidth]{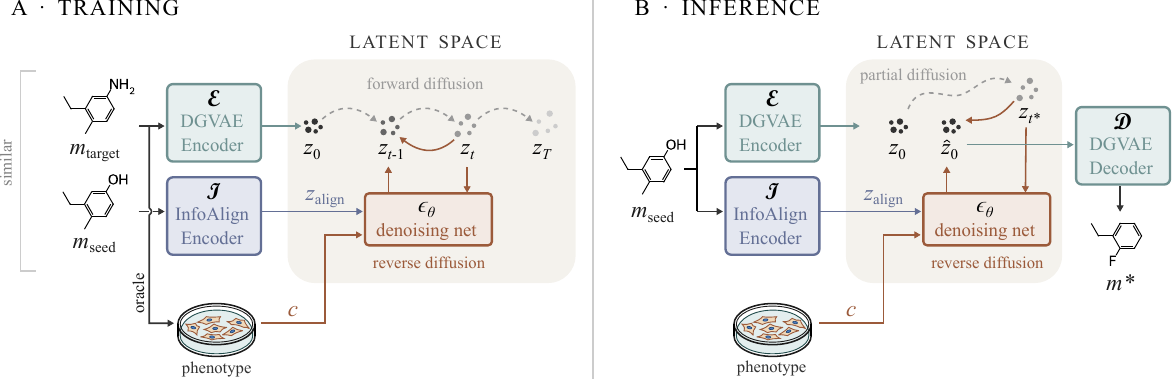}
    \caption{\textbf{Overview of PhAME.} \textit{(Training, left)} A frozen DGVAE encoder $\mathcal{E}$ maps a target molecule to a latent $z_0$, which is noised along a forward diffusion, and a phenotype-aware encoding $z_\text{align}$ of a seed molecule along with a target phenotypic signature $c$ is used to train a denoising network. \textit{(Inference, right)} A molecule is encoded into the latent space and partially noised; then, the denoising network conditioned on phenotype-aware encoding $z_\text{align}$ of the same molecule and phenotypic signature $c$ is used to generate a molecular analog, which is decoded by the DGVAE decoder $\mathcal{D}$.}
    \label{fig:overview}
\end{figure}

\subsection{PhAME}
\label{sec:model}

PhAME (overview in Figure~\ref{fig:overview}) operates in the latent space of DGVAE~\citep{Nguyen2025VAE}, a pretrained VAE that encodes molecular graphs and decodes SMILES strings. The graph encoder yields latent codes $z_0 = \mathcal{E}(m)$ that are invariant to SMILES string ordering, while the SMILES decoder $\mathcal{D}$ maps back to valid molecules. The encoder and decoder are frozen; we only train the diffusion model on top of this latent space, following the formulation in Section~\ref{sec:background}.

The denoising network $\epsilon_\theta$ is conditioned on two signals in addition to the timestep $t$: a target signature $c \in \mathbb{R}^K$ and a molecular embedding $z_{\text{align}}$ that anchors generation to the seed molecule's neighborhood. We obtain $z_{\text{align}}$ from InfoAlign~\citep{liu2025learning}, which contains an information bottleneck that aligns molecular structure with Cell Painting morphological profiles~\citep{bray2016cell} and L1000 gene expression profiles~\citep{Subramanian_L1000}. Since InfoAlign retains structural information that is predictive of cellular phenotype, $z_{\text{align}}$ encodes phenotypically relevant molecular similarity rather than raw fingerprint distance. The embedding is computed once per seed and held fixed during denoising.

\paragraph{Training.}
The model is trained on molecular pairs constructed as described in Section~\ref{sec:dataset}: for each pair, $z_0 = \mathcal{E}(m_{\text{target}})$ is the DGVAE latent of the target molecule with signature $c$, and $z_{\text{align}}$ is the InfoAlign embedding of the corresponding seed molecule. We use a cosine noise schedule~\citep{nichol2021improved} and train with the denoising objective from Section~\ref{sec:background}, extended to the dual-conditioned network: $\mathcal{L}_{\text{diff}} = \mathbb{E}_{z_0,\epsilon,t}\bigl[\|\epsilon - \epsilon_\theta(z_t, t, c, z_{\text{align}})\|^2\bigr]$. Since $\mathcal{L}_{\text{diff}}$ alone does not constrain where in latent space the denoised sample $\hat{z}_0$ lands, we add an alignment loss
\begin{equation}
\label{eq:L_align}
\mathcal{L}_{\text{align}} =
\max\!\left(0,\; \tau - \cos(\hat{z}_0,\, \psi(z_{\text{align}}))\right),
\end{equation}
with margin $\tau{=}0.8$, penalizing predictions that drift too far from the seed. Function $\psi$ is a two-layer MLP that projects the compound representation into the VAE latent space. The full objective is $\mathcal{L} = \mathcal{L}_{\text{diff}} + \gamma\, \mathcal{L}_{\text{align}}$. During training, $c$ and $z_{\text{align}}$ are each independently replaced with a null token $\varnothing$ with probability $p_{\text{uncond}}=0.1$, so the network sees all four conditioning combinations.

\paragraph{Compositional guidance.}
Standard CFG (Section~\ref{sec:background}) uses a single scale $w$ for all conditioning, conflating signature guidance with structural anchoring. We decompose guidance into two terms with separate scales $w_c$ and $w_a$. Let $\Delta_c = \epsilon_{\theta}(\hat{z}_t, t, c, \varnothing) - \epsilon_{\theta}(\hat{z}_t, t, \varnothing, \varnothing)$ and $\Delta_a = \epsilon_{\theta}(\hat{z}_t, t, \varnothing, z_{\text{align}}) - \epsilon_{\theta}(\hat{z}_t, t, \varnothing, \varnothing)$ denote the signature and alignment directions. The guided prediction is
\begin{equation}
\label{eq:multi-cfg}
\tilde{\epsilon}_{\theta}(\hat{z}_t, t, c, z_{\text{align}}) =
\epsilon_{\theta}(\hat{z}_t, t, \varnothing, \varnothing)
+ w_c \,\Delta_c + w_a \,\Delta_a.
\end{equation}
Here $w_c$ controls how strongly generation is pushed toward the target signature, while $w_a$ controls how tightly it stays near the seed structure.

\paragraph{Inference.}
PhAME can run in two modes. In \textit{molecular optimization}, we encode the seed molecule $z_0 = \mathcal{E}(m)$, run $t^* < T$ forward noising steps, and denoise with Eq.~\ref{eq:multi-cfg}; the noise level $t^*$ controls the exploration radius around the seed, while $w_c$ and $w_a$ independently steer signature and structural guidance. In \textit{de novo generation}, we sample $\hat{z}_T \sim \mathcal{N}(0, I)$ and denoise from pure noise with $\gamma = 0$ and $w_a = 0$, reducing the model to a signature-conditioned generator with no structural anchor.

\subsection{Dataset}
\label{sec:dataset}

For each task, we partition the dataset according to a predefined property threshold (e.g., solubility or docking score), yielding two disjoint subsets corresponding to molecules with property values above and below the threshold.

To construct the training data, molecules from the two subsets are paired by selecting the most structurally similar counterparts across the threshold. Specifically, for each molecule in one subset, we identify the nearest neighbor in the opposite subset based on molecular similarity and form a training pair. Structural similarity is computed using the Tanimoto similarity between Morgan fingerprints. This pairing strategy provides supervision signals that encourage the model to learn transformations that move molecules across the threshold while preserving structural similarity to the source structure.

\section{Results}

We evaluate the proposed approach across a range of benchmark and real-world molecular design tasks. In Section~\ref{sec:logP}, we examine the contribution of individual model components within a controlled logP optimization setting, and benchmark our method against state-of-the-art approaches for molecular editing. In Section~\ref{sec:docking}, we evaluate the model’s ability to generate novel compounds with high predicted binding affinity while preserving drug-likeness and synthetic accessibility. Finally, in Sections~\ref{sec:genes} and~\ref{sec:phenotype}, we demonstrate the model’s capability for multimodal conditioning, enabling molecular generation guided by gene expression profiles and phenotypic signals.

\subsection{Controlled Property Editing on logP}
\label{sec:logP}

We first evaluate PhAME on a controlled logP optimization benchmark against state-of-the-art molecular editing methods. Moreover, we show that independently tuning phenotype guidance ($w_c$) and structural guidance ($w_a$) provides explicit control over the trade-off between property optimization and structural preservation.

\paragraph{Setup.}

We evaluate on the MOSES benchmark~\cite{polykovskiy2020molecular} using the standard train/test split. Following Section~\ref{sec:dataset}, we partition molecules according to the median logP of the training set to define the optimization direction. The task is bidirectional: high-logP molecules are optimized downward and low-logP molecules upward, providing a controlled setting for assessing property-guided generation. 
\paragraph{Baselines.}

Given that oracle-guided optimization is rarely feasible in real-world drug discovery due to prohibitive costs and limited assay availability, our benchmark evaluates methods that operate without access to optimization oracles. We compare against state-of-the-art molecular editing and generation methods, including Mol-CycleGAN~\cite{maziarka2020mol}, DiGress~\cite{vignac2023digress}, and PURE~\cite{gupta2025pure}. To contextualize task difficulty and dataset structure, we additionally include a set of simple retrieval-based baselines. \emph{\mbox{Random target class}} retrieves a randomly selected molecule from the target partition. \emph{\mbox{Nearest target class}} retrieves the closest molecule within the target partition, approximating an ideal similarity-preserving transformation. \emph{VAE reconstruction} reconstructs the input molecule without explicitly optimizing the target property. Full details are provided in Appendix~\ref{sec:app_logP_optimization}.

\paragraph{Metrics.} 

We evaluate generated molecules using four metrics. \emph{Nov} (novelty) is the fraction of molecules not present in the training set. \emph{Tgt} (target success) measures the fraction of molecules that reach the desired property region. \emph{Sim} denotes the Tanimoto similarity to the seed molecule. \emph{NTS} is defined as the product of \emph{Nov}, \emph{Tgt}, and \emph{Sim}, capturing the ability to generate novel molecules that satisfy the target objective while remaining close to the seed structure. Additional quality metrics of generated molecules are reported in Appendix~\ref{sec:app_logP_optimization}.
\begin{wrapfigure}{r}{0.5\textwidth}
\vspace{-\intextsep}
\centering
\scriptsize
\setlength{\tabcolsep}{1.5pt}
\renewcommand{\std}[1]{{\tiny\,$\pm$#1}}
\renewcommand{\arraystretch}{1.1}
\vspace{1.0cm}
\captionof{table}{\textbf{LogP optimization with similarity preservation to the seed molecule.} While competing methods typically optimize either target success (\emph{Tgt}) or structural similarity (\emph{Sim}), PhAME achieves the strongest overall performance on the metric \emph{NTS}. We report mean${}_{\pm\text{std}}$ over 5 runs.}
\resizebox{\linewidth}{!}{
\begin{tabular}{lccc|c}
\toprule
Method & Nov $\uparrow$ & Tgt $\uparrow$ & Sim $\uparrow$ & NTS $\uparrow$ \\
\midrule
Random target class  
& 0.00\std{0.00} 
& 1.00\std{0.00} 
& 0.14\std{0.00} 
& 0.00\std{0.00} \\
Nearest target class
& 0.00\std{0.00} 
& 1.00\std{0.00} 
& 0.60\std{0.00} 
& 0.00\std{0.00} \\
VAE reconstruction 
& 0.99\std{0.00} 
& 0.19\std{0.00} 
& 0.59\std{0.00} 
& 0.11\std{0.00} \\
\midrule
Mol-CycleGAN 
& 0.99\std{0.00}
& 0.51\std{0.04} 
& 0.15\std{0.00} 
& 0.08\std{0.01} \\
DiGress & 1.00\std{0.00} 
& 1.00\std{0.00} 
& 0.13\std{0.00} 
& 0.13\std{0.00} \\
PURE & 1.00\std{0.00} 
& 0.45\std{0.00} 
& 0.58\std{0.00} 
& 0.26\std{0.00} \\
\midrule

PhAME w/o $\mathcal{L}_{\text{align}}$ 
& 0.99\std{0.00} 
& 0.81\std{0.01} 
& 0.35\std{0.00} 
& 0.28\std{0.00} \\

PhAME (ours)
& 0.99\std{0.00}
& 0.78\std{0.03}
& 0.39\std{0.02}
& \textbf{0.30\std{0.01}} \\
\bottomrule
\end{tabular}}
\label{tab:logp_benchmark}
\vspace{-1.0cm}
\end{wrapfigure}

\begin{wrapfigure}{r}{0.5\textwidth}
\vspace{-\intextsep}

\vspace{0.1cm}

\includegraphics[width=\linewidth]{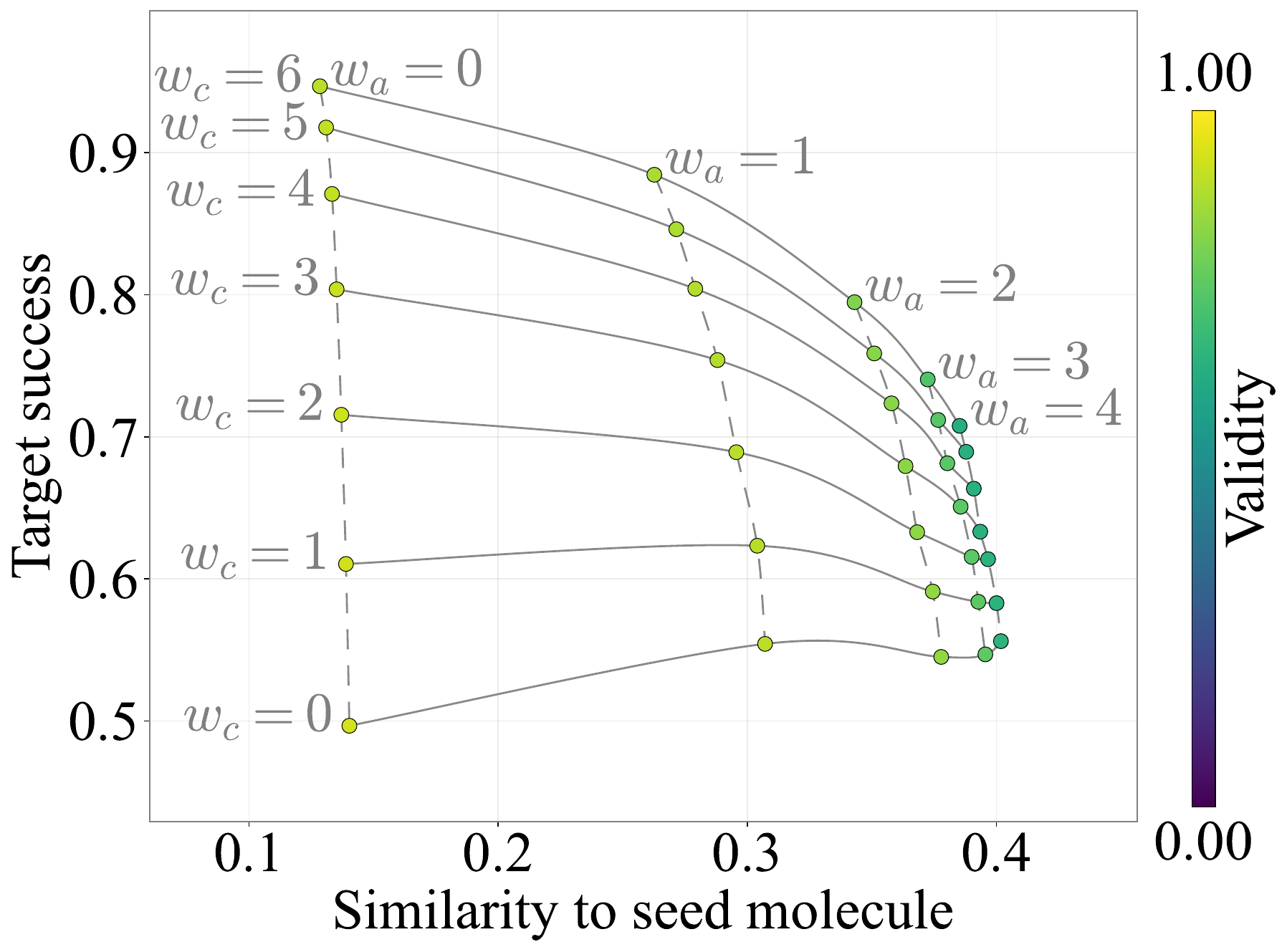}
\captionof{figure}{\textbf{Influence of guidance scales}. Trade-off between target success and similarity under varying guidance strengths. Increasing property guidance ($w_c$) improves target success, while increasing alignment guidance ($w_a$) enhances similarity to the seed molecule.}
\label{fig:guidance_optimization}

\vspace{-0.5cm}

\end{wrapfigure}

\paragraph{Results.}
 Table~\ref{tab:logp_benchmark} highlights a clear trade-off between structural preservation and property optimization among the baselines. Mol-CycleGAN struggles to achieve both high target success and structural similarity. DiGress attains perfect target success but sacrifices structural alignment, whereas PURE exhibits the opposite behavior, preserving similarity while underperforming on the optimization objective. In contrast, PhAME ($w_c=6, w_a=3$) performs strongly across all metrics (\emph{Nov}, \emph{Tgt}, and \emph{Sim}) and achieves the best overall \emph{NTS}, demonstrating a superior balance between novelty, target satisfaction, and structural preservation. Incorporating the alignment loss $\mathcal{L}_{\text{align}}$ further improves structural preservation relative to the variant without $\mathcal{L}_{\text{align}}$, increasing similarity while maintaining competitive target success and resulting in a higher \emph{NTS} score.

Notably, PhAME is highly steerable, as illustrated in Figure~\ref{fig:guidance_optimization} and detailed in Table~\ref{tab:app_phame_ablation} in Appendix~\ref{sec:app_logP_optimization}. This flexibility allows the model to match or approach the individual strengths of specialized methods, such as \emph{Tgt} of DiGress or \emph{Sim} of PURE. The compositional-guidance mechanism enables independent control over the optimization trade-off. Increasing property guidance ($w_c$) consistently improves target success, with diminishing returns at larger values and minimal effect on similarity. In contrast, increasing alignment guidance ($w_a$) improves similarity but weakens property optimization. These gains also saturate at larger $w_a$, while validity decreases.

Overall, the results show that jointly leveraging both guidance signals is essential, as neither property nor structural guidance alone is sufficient. Their combination enables effective control over the trade-off between target optimization and similarity to the seed molecule.
\subsection{Docking Score Optimization}
\label{sec:docking}

We assess PhAME's ability to generate \emph{novel} high-affinity binders against established protein targets. Unlike the editing setting elsewhere in this paper, this benchmark has no seed compound: we sample from the DGVAE prior and denoise the random latents under PhAME's compositional guidance, with binding affinity scored by QuickVina2~\citep{alhossary2015fast}. To push the model toward increasingly potent regions of chemical space, training follows a three-stage curriculum: each stage fine-tunes the denoising network on ZINC250k pairs split at a stricter binder/non-binder docking-score threshold than the previous, progressively sharpening the property gradient the model must learn to follow.

\paragraph{Setup.} Following Lee et al.~\citep{lee2023exploring}, we train on ZINC250k~\citep{irwin2005zinc} and generate 3{,}000 candidates each for the five protein targets PARP1, FA7, 5HT1B, BRAF, and JAK2, scoring them with QuickVina2 alongside QED and SA. We report two metrics from Yang et al.~\citep{yang2021hit} with the novelty constraint of Lee et al.~\citep{lee2023exploring}: \emph{novel top 5\% docking score} (\emph{Dock.}, kcal/mol), the average docking score over the top 5\% of generations satisfying $\text{QED}{>}0.5$, $\text{SA}{<}5$, and max Tanimoto-to-train ${<}0.4$; and \emph{novel hit ratio} (\emph{Hit}), the fraction of generations passing those filters and achieving a target-specific docking threshold. We compare against the seven baselines reported by Lee et al.~\citep{lee2023exploring}, a \emph{Random VAE} sanity check that decodes Gaussian samples from the DGVAE prior without guidance, and an upper-reference computed on the top 5\% of ZINC250k itself. Curriculum schedule, per-target thresholds, hyperparameters, and the Random VAE definition are in Appendix~\ref{sec:app_docking_score_optimization}.

\begin{table*}[t]
\centering
\caption{\textbf{Novel top 5\% docking score (\emph{Dock.}) and novel hit ratio (\emph{Hit}).} Docking scores are reported in kcal/mol and hit ratios in \%. PhAME achieves the best docking scores for all targets and the highest hit ratio for 4 out of 5 proteins. Cells report mean${}_{\pm\text{std}}$ over 5 runs; ZINC250k is a top-5\% reference and has no std. Best results are in bold and second-best underlined.}
\label{tab:novel_docking_hitratio}
\scriptsize
\renewcommand{\std}[1]{{\tiny\,$\pm$#1}}
\setlength{\tabcolsep}{2pt}
\renewcommand{\arraystretch}{1.1}

\resizebox{\textwidth}{!}{%
\begin{tabular}{l rr rr rr rr rr}
\toprule
& \multicolumn{2}{c}{PARP1}
& \multicolumn{2}{c}{FA7}
& \multicolumn{2}{c}{5HT1B}
& \multicolumn{2}{c}{BRAF}
& \multicolumn{2}{c}{JAK2} \\
\cmidrule(lr){2-3}
\cmidrule(lr){4-5}
\cmidrule(lr){6-7}
\cmidrule(lr){8-9}
\cmidrule(lr){10-11}
Method
& \multicolumn{1}{c}{Dock.$\downarrow$} & \multicolumn{1}{c}{Hit$\uparrow$}
& \multicolumn{1}{c}{Dock.$\downarrow$} & \multicolumn{1}{c}{Hit$\uparrow$}
& \multicolumn{1}{c}{Dock.$\downarrow$} & \multicolumn{1}{c}{Hit$\uparrow$}
& \multicolumn{1}{c}{Dock.$\downarrow$} & \multicolumn{1}{c}{Hit$\uparrow$}
& \multicolumn{1}{c}{Dock.$\downarrow$} & \multicolumn{1}{c}{Hit$\uparrow$} \\
\midrule

ZINC250k
& \multicolumn{1}{c}{-10.59} & \multicolumn{1}{c}{5.66}
& \multicolumn{1}{c}{-8.48}  & \multicolumn{1}{c}{1.61}
& \multicolumn{1}{c}{-10.51} & \multicolumn{1}{c}{27.36}
& \multicolumn{1}{c}{-10.28} & \multicolumn{1}{c}{1.62}
& \multicolumn{1}{c}{-9.75}  & \multicolumn{1}{c}{7.77} \\

\midrule

REINVENT
& -8.70 \std{0.52} & 0.48 \std{0.34}
& -7.21 \std{0.26} & 0.21 \std{0.08}
& -8.77 \std{0.32} & 2.45 \std{0.56}
& -8.39 \std{0.40} & 0.13 \std{0.09}
& -8.17 \std{0.28} & 0.61 \std{0.17} \\

MORLD
& -7.53 \std{0.26} & 0.05 \std{0.05}
& -6.26 \std{0.17} & 0.01 \std{0.01}
& -7.87 \std{0.65} & 0.88 \std{0.74}
& -8.04 \std{0.34} & 0.05 \std{0.04}
& -7.82 \std{0.13} & 0.23 \std{0.12} \\

HierVAE
& -9.49 \std{0.28} & 0.55 \std{0.21}
& -6.81 \std{0.27} & 0.01 \std{0.01}
& -8.08 \std{0.25} & 0.51 \std{0.28}
& -8.98 \std{0.53} & 0.21 \std{0.22}
& -8.29 \std{0.37} & 0.23 \std{0.13} \\

FREED
& -10.43 \std{0.18} & 3.63 \std{0.96}
& -8.30 \std{0.09} & 1.11 \std{0.21}
& -10.43 \std{0.33} & 10.19 \std{3.31}
& -10.33 \std{0.16} & 2.07 \std{0.63}
& -9.62 \std{0.10} & 4.52 \std{0.67} \\

FREED-QS
& -10.58 \std{0.10} & 4.63 \std{0.73}
& \underline{-8.38 \std{0.04}} & \underline{1.33 \std{0.11}}
& -10.71 \std{0.18} & \underline{16.77 \std{0.90}}
& -10.56 \std{0.08} & 2.94 \std{0.36}
& -9.74 \std{0.02} & 5.80 \std{0.30} \\

GDSS
& -9.97 \std{0.03} & 1.93 \std{0.21}
& -7.78 \std{0.04} & 0.37 \std{0.10}
& -9.46 \std{0.10} & 4.67 \std{0.31}
& -9.22 \std{0.07} & 0.17 \std{0.13}
& -8.93 \std{0.09} & 1.17 \std{0.28} \\

MOOD
& \underline{-10.87 \std{0.11}} & \underline{7.02 \std{0.43}}
& -8.16 \std{0.07} & 0.73 \std{0.14}
& \underline{-11.15 \std{0.04}} & \textbf{18.67 \std{0.42}}
& \underline{-11.06 \std{0.03}} & \underline{5.24 \std{0.29}}
& \underline{-10.15 \std{0.06}} & \underline{9.20 \std{0.52}} \\

\midrule

Random VAE
& -8.51 \std{0.04} & 0.10 \std{0.04}
& -7.17 \std{0.03} & 0.04 \std{0.03}
& -8.48 \std{0.07} & 1.43 \std{0.14}
& -8.58 \std{0.04} & 0.03 \std{0.03}
& -8.28 \std{0.05} & 0.33 \std{0.06} \\

PhAME (ours)
& \textbf{-11.17 \std{0.04}} & \textbf{11.04 \std{0.82}}
& \textbf{-9.09 \std{0.10}} & \textbf{5.75 \std{0.76}}
& \textbf{-11.29 \std{0.12}} & 14.62 \std{1.15}
& \textbf{-11.12 \std{0.13}} & \textbf{6.36 \std{0.64}} 
& \textbf{-10.64 \std{0.05}} & \textbf{14.84 \std{1.73}} \\

\bottomrule
\end{tabular}%
}
\end{table*}

\begin{figure}[bt]
    \centering
    \includegraphics[width=\linewidth]{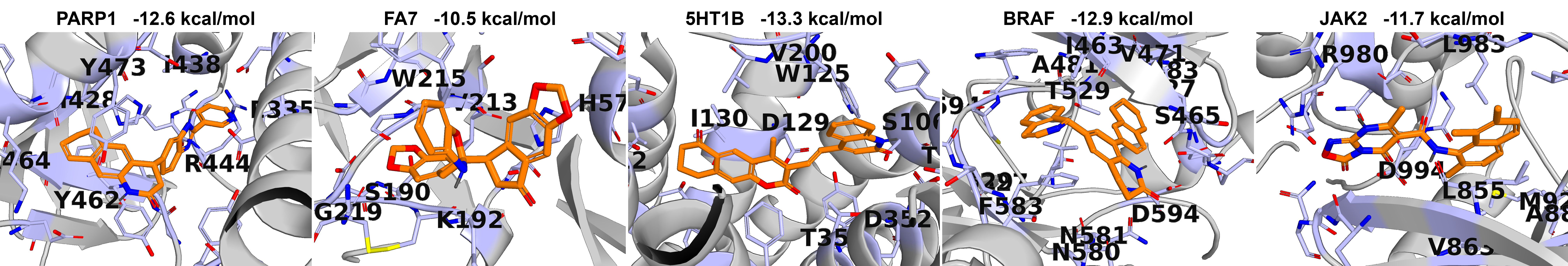}
    \caption{\textbf{Top-scoring docked poses of PhAME-generated molecules across five protein targets.} For each target, we visualize the highest-affinity molecule generated under docking-score guidance. The protein backbone is shown as cartoon; binding-pocket residues (any heavy-atom contact within 3.5\,\AA{} of the ligand) are drawn as light-blue sticks; the docked ligand is in orange.}
    \label{fig:docked-poses}
    \vspace{-1.3em}
\end{figure}
\begin{wrapfigure}{r}{0.5\textwidth}
\centering
\vspace{-0.4cm}

{\tiny
\setlength{\tabcolsep}{2pt}
\captionof{table}{\textbf{Average molecular quality metrics across targets.} PhAME achieves the strongest overall molecular quality among gene-conditioned generation baselines. Cells report mean${}_{\pm\text{std}}$ over targets.}
\label{tab:stats_genes}
\renewcommand{\std}[1]{{\tiny\,$\pm$#1}}
\resizebox{\linewidth}{!}{
\begin{tabular}{lcccc}
\toprule
Method & QED $\uparrow$ & SA $\downarrow$ & Uniqueness $\uparrow$ & Novelty $\uparrow$ \\
\midrule
TRIOMPHE & 0.39\std{0.03} & 6.02\std{0.29} &\textbf{1.00\std{0.00}} & \textbf{1.00 \std{0.00}} \\
GxVAEs & 0.59\std{0.04} & 3.42\std{0.32} & 0.86\std{0.09} & 0.45\std{0.20} \\
SmilesGEN  & 0.60\std{0.02} & 3.48\std{0.18} & 0.94\std{0.03} & 0.16 \std{0.14} \\
PhAME (ours) & \textbf{0.71\std{0.03}} & \textbf{2.94\std{0.17}} & \textbf{1.00\std{0.00}} & \textbf{1.00\std{0.00}} \\
\bottomrule
\end{tabular}}}
\vspace{-0.4cm}

\end{wrapfigure}

\paragraph{Results.} Table~\ref{tab:novel_docking_hitratio} shows PhAME setting a new state of the art across the benchmark. PhAME achieves the best \emph{Dock.}~score on \emph{all five targets} and is the only method whose top-5\% generations surpass the top-5\% of ZINC250k uniformly: PhAME's 3{,}000 outputs contain a head that is more potent than the strongest 12{,}500 molecules of its own training set. On novel hit ratio, PhAME leads on four of five targets, with relative gains of $+332\%$ over the next-best method on FA7, $+61\%$ on JAK2, and $+57\%$ on PARP1.
The single regression is 5HT1B, where PhAME (14.6\%) trails MOOD (18.7\%) and FREED-QS (16.8\%). On this target, however, ZINC250k itself reaches a 27.4\% novel hit rate, an order of magnitude above the other four, meaning the benchmark is dominated by the $\text{Tanimoto}{<}0.4$ novelty filter rather than by generative capacity: high-affinity 5HT1B chemotypes are densely populated in ZINC and any sufficiently active candidate is likely to share a scaffold with the training set. The gap between PhAME and MOOD here is small relative to the gaps PhAME opens on the other four targets. 

To attribute the gain to the three-stage curriculum rather than to the DGVAE latent space or CFG alone, we ablate the curriculum on FA7 in Appendix~\ref{sec:app_docking_score_optimization}: collapsing PhAME to a single stage drops the novel hit ratio markedly, confirming that the progressive sharpening of the binder/non-binder threshold is what unlocks the improvement. Top-scoring generated hits are visualized in Figure~\ref{fig:docked-poses}.

\subsection{Transcriptomic Guided Generation}
\label{sec:genes}

\begin{wrapfigure}{r}{0.5\textwidth}
\centering
\vspace{-\intextsep}

\vspace{-2.8pt}
\includegraphics[width=\linewidth]{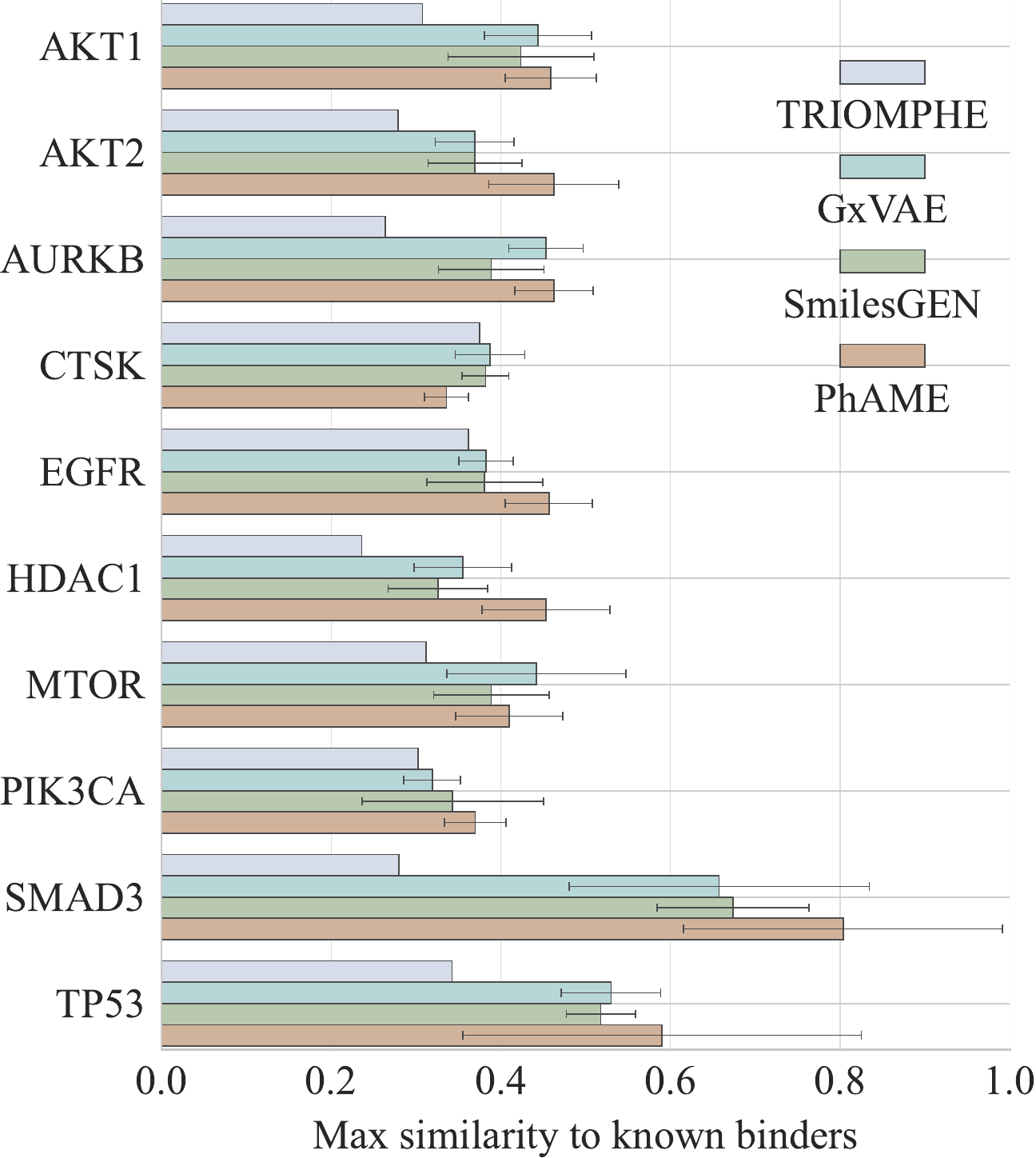}
\caption{\textbf{Recovery of known binder motifs}. Maximum Tanimoto similarity to known binders for each target. PhAME generates compounds most similar to known binders for 8 out of 10 targets. Bars show mean $\pm$ std over 5 runs.}
\label{fig:sim_to_binders}
\vspace{0.0cm}

\includegraphics[width=1\linewidth]{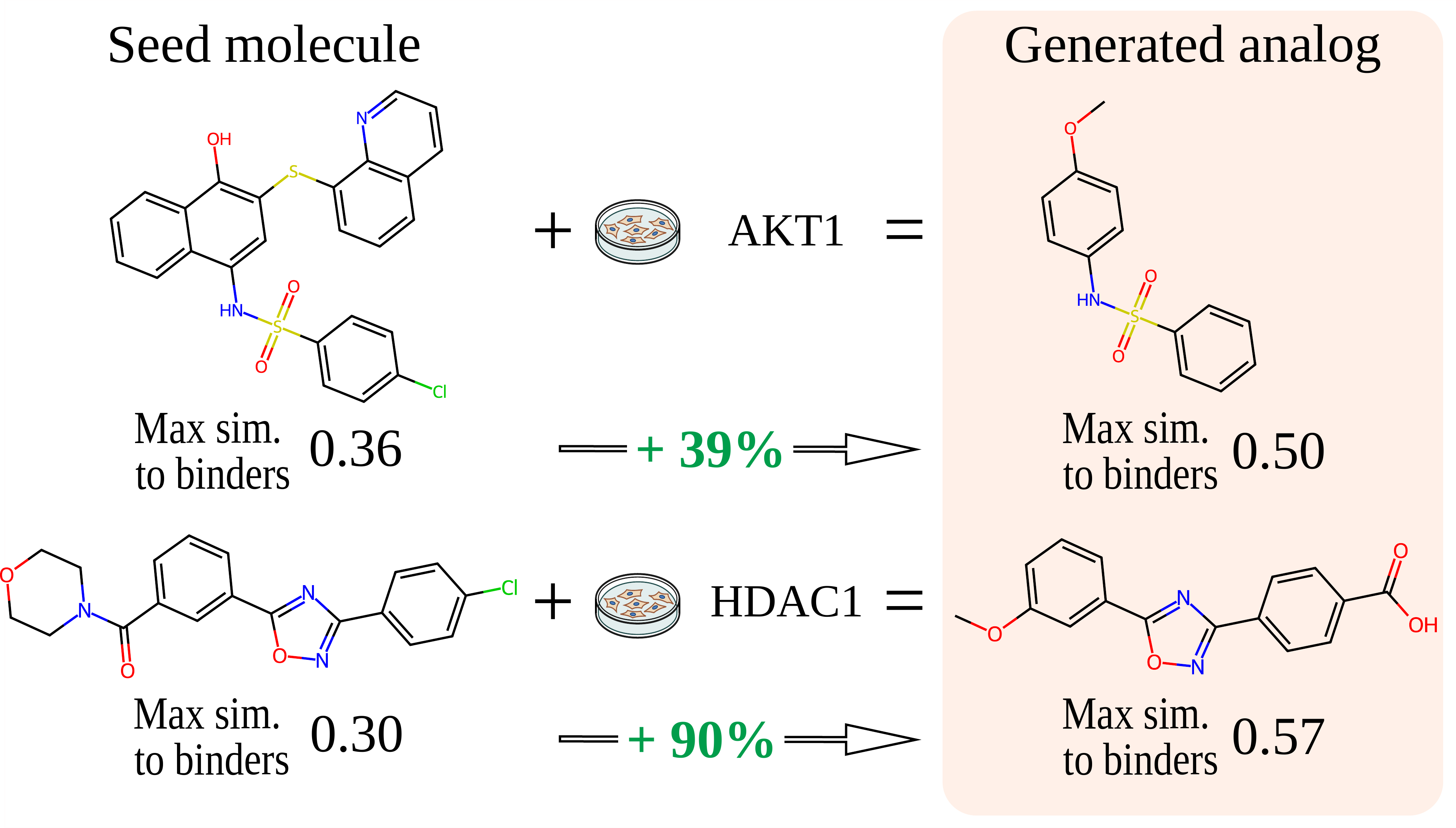}
\caption{\textbf{Gene-driven recovery of known binders.} PhAME generates analogs with substantially increased similarity to known binders using only target-specific knockdown profiles. For AKT1, the maximum Tanimoto similarity to known binders improves by 39\%, while for HDAC1 the improvement reaches 90\%.}
\label{fig:gene_editing}
\vspace{-1.8cm}

\end{wrapfigure}

Complementary to the structure-based docking
task, we evaluate PhAME on generating compounds conditioned on target-specific transcriptomic perturbations. This setting assesses whether the model can translate gene expression signatures into chemically plausible and target-relevant molecules, without relying on explicit structural or docking information.

\paragraph{Setup.}

We follow the evaluation protocol of Li \& Yamanishi~\citep{Li_2024_gxvaes,Hui_2025_smilesgen}, where models are assessed on their ability to generate compounds conditioned on target-specific transcriptomic perturbations. Models are trained on paired transcriptomic–molecular data and evaluated against known ligands. For each target, we generate 100 molecules and report the maximum Tanimoto similarity of generated compounds to known ligands. Knockdown profiles are used to generate inhibitors for \textit{AKT1}, \textit{AKT2}, \textit{AURKB}, \textit{CTSK}, \textit{EGFR}, \textit{HDAC1}, \textit{MTOR}, and \textit{PIK3CA}, while overexpression profiles are used for activators of \textit{SMAD3} and \textit{TP53}. Transcriptomic signatures are obtained from LINCS~\citep{duan2014lincs} using the \textit{MCF7} cell line.
While largely following Li \& Yamanishi~\citep{Li_2024_gxvaes}, we identify data leakage due to overlap between training and evaluation molecules and remove the overlapping compounds, reducing the training set from 13,755 to 13,432 ($<3\%$) (details in Appendix~\ref{sec:app_transcriptomic_guided_generation}).

\paragraph{Results.}

Table~\ref{tab:stats_genes} shows that PhAME outperforms prior gene-conditioned models, achieving the highest drug-likeness (QED) and lowest synthetic accessibility (SA), while maintaining perfect uniqueness and novelty. In contrast, SmilesGEN exhibits low novelty, suggesting memorization of training molecules. Figure~\ref{fig:sim_to_binders} shows that PhAME generates compounds structurally similar to known binders, achieving the highest Tanimoto similarity for 8 of 10 targets. 

Figure~\ref{fig:gene_editing} demonstrates PhAME’s editing capability. Generation is initialized from the nearest training molecule in gene-expression space (cosine similarity). The left panel shows the starting compounds, and the right panel edited counterparts conditioned on target-specific expression profiles. In both cases, PhAME preserves the core molecular structure while increasing similarity to known binders, using only target knockdown profiles. For AKT1, PhAME improves similarity to known binders by 39\%, while for HDAC1 the improvement reaches 90\%.

\subsection{Cell Phenotype Guided Generation}
\label{sec:phenotype}

We close with the most demanding setting in this paper: generating \emph{de novo} molecules conditioned on a Cell Painting morphology centroid corresponding to a target Mode of Action (MoA). There is no synthesizable oracle that can rate a generated compound on its true cellular response, so we evaluate via a structural proxy: do the generations sit closer to the reference compounds of target MoA than to those of any other MoA, across multiple representation spaces?

\paragraph{Setup.} We use the JUMP Cell Painting dataset~\citep{chandrasekaran2023align}, for evaluation we intersected it with ChEMBL2K~\citep{gaulton2012chembl} and cross-referenced with the Broad Drug Repurposing Hub~\citep{corsel2017hub} to obtain MoA labels, restricted to seven balanced MoA classes (8 reference compounds each). PhAME is run in its \textit{de novo} mode (Section~\ref{sec:model}): we condition on the CellProfiler~\citep{carpenter2006cellprofiler} morphological centroid of each target MoA and sample 400 candidates per class (5 seeds $\times$ 80 valid samples). Dataset construction, feature provenance, contrastive-pairing thresholds, and inference hyperparameters are in Appendix~\ref{sec:app_cell_phenotype_guided_generation}; baselines (DGVAE, CPMolGAN~\citep{zapata2023cell}, GFlowNet~\citep{lu2024cell}) follow their original protocols.

\paragraph{Metrics.}
\label{MetricsPhenotype}
We score each generation against the seven MoA classes in three representation spaces with complementary inductive biases: ECFP (raw structural similarity), CLOOME embeddings~\citep{sanchez2023cloome} (jointly trained on molecule--cell pairs), and InfoAlign embeddings (information bottleneck over morphology and L1000). We report: \emph{Top-1 Cluster Accuracy} (does the highest-scoring MoA class match the target?), \emph{MoA Retrieval Rate @$k$} (does the target rank in the top $k$ classes by average similarity to its $n{=}3$ nearest reference compounds?), and \emph{kNN Accuracy @$k$} (majority-vote MoA among the $k$ nearest references). The chance baseline for Top-1 Cluster Accuracy is $1/7 \approx 14.3\%$. The formal definitions of these metrics are given in Appendix~\ref{sec:app_cell_phenotype_guided_generation}.

\begin{table}[t]
\centering
\caption{\textbf{MoA classification of \textit{de novo} generated molecules evaluated in the ECFP structural embedding space.} Values are percentages (mean${}_{\pm\text{std}}$ over 5 seeds). Best in bold, second-best underlined. PhAME fundamentally outperforms competing methods, uniquely demonstrating superior results on all evaluated metrics.}
\label{tab:cell}
\scriptsize
\renewcommand{\std}[1]{{\tiny\,$\pm$#1}}
\setlength{\tabcolsep}{6pt}
\renewcommand{\arraystretch}{1.1}

\resizebox{\textwidth}{!}{%
\begin{tabular}{ll c ccc ccc}
\toprule
& \multicolumn{1}{c}{\textbf{Top-1}}
& \multicolumn{3}{c}{\textbf{MoA Retrieval Rate @$k$}~$\uparrow$}
& \multicolumn{3}{c}{\textbf{kNN Accuracy @$k$}~$\uparrow$} \\
\cmidrule(lr){2-2} \cmidrule(lr){3-5} \cmidrule(lr){6-8}
 \textbf{Model}
& \textbf{Cluster Acc}~$\uparrow$
& \textbf{@1} & \textbf{@2} & \textbf{@3}
& \textbf{@1} & \textbf{@3} & \textbf{@5} \\
\midrule
 DGVAE        & 14.0 \std{0.9} & 14.2 \std{2.0} & 28.5 \std{1.1} & 42.7 \std{2.1} & \underline{14.5 \std{0.6}} & \underline{14.6 \std{1.0}} & 13.7 \std{1.6} \\
 CPMolGAN     & 14.3 \std{1.0} & 13.4 \std{1.3} & 28.8 \std{2.4} & 42.0 \std{0.9} & 12.9 \std{1.2} & 13.5 \std{1.4} & 14.1 \std{1.4} \\
 GFlowNet     & \underline{17.0 \std{1.8}} & \underline{15.1 \std{1.3}} & \underline{29.0 \std{1.6}} & \underline{46.1 \std{2.1}} & 12.7 \std{1.1} & 14.4 \std{1.3} & \underline{15.3 \std{1.6}} \\
 PhAME (ours) & \textbf{27.2 \std{1.8}} & \textbf{23.0 \std{1.6}} & \textbf{40.4 \std{2.3}} & \textbf{56.2 \std{2.9}} & \textbf{20.6 \std{1.2}} & \textbf{21.2 \std{1.6}} & \textbf{23.0 \std{2.2}} \\
\bottomrule
\end{tabular}%
}
\vspace{-2em}
\end{table}

\begin{wrapfigure}{r}{0.5\textwidth}
\vspace{-0.5cm}
\includegraphics[width=\linewidth]{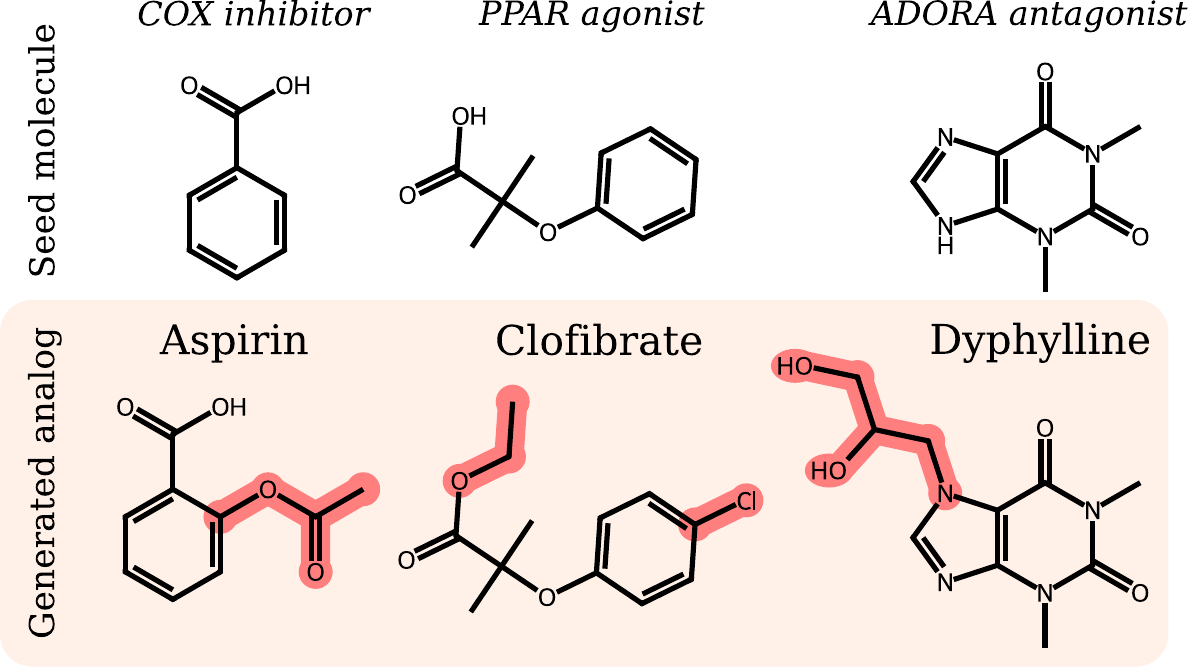}
\caption{\textbf{Phenotype-driven recovery of known drugs.} PhAME, run in editing mode and seeded with a structurally similar hit, recovers analogs: aspirin (\emph{cyclooxygenase inhibitor}, left), clofibrate (\emph{PPAR receptor agonist}, middle) and dyphylline (\emph{adenosine receptor antagonist}, right) when conditioned on the corresponding MoA centroid.}
\label{fig:h2l_cell}
\vspace{-0.5cm}
\end{wrapfigure}

\paragraph{Results.} Table~\ref{tab:cell} reveals a clean separation for evaluation on ECFP embedding space. DGVAE and CPMolGAN sit at the chance baseline ($\sim$14.3\% Top-1 Cluster Accuracy), recovering essentially no MoA information. GFlowNet shows only a marginal improvement over these baselines. In contrast, PhAME achieves a $+60\%$ relative gain over GFlowNet on Top-1 Cluster Accuracy ($0.272$ vs.\ $0.170$), with the same pattern repeating across MoA Retrieval Rate @$k$ and kNN Accuracy @$k$. 

Crucially, extended results in Appendix~\ref{sec:app_cell_phenotype_guided_generation} (Table~\ref{tab:app_cell}) confirm that PhAME is the \emph{only} method to clear the chance baseline across all three spaces (ECFP, CLOOME, InfoAlign). It significantly outperforms GFlowNet in InfoAlign (where the biological signal is richest) and recovers from a minor deficit in CLOOME Top-1 accuracy by dominating deeper retrieval metrics (MoA @1/2/3, kNN @1/3/5). Because these advantages hold consistently across structurally diverse representations, the gains reflect genuine MoA-aware chemistry rather than embedding-specific overfitting.

\paragraph{From de novo to drug recovery.}
The same model can be run in editing mode (Section~\ref{sec:model}) to recover known therapeutics: seeding with a structurally similar hit and conditioning on the target MoA, PhAME reconstructs aspirin, clofibrate and dyphylline from MoA-only guidance (Figure~\ref{fig:h2l_cell}). Additional details are in Appendix~\ref{app:cell_phenotype_optimization}.

\section{Conclusions}
In this work, we introduced PhAME which combines discrete chemical graphs with high-dimensional biological signatures to formulate molecular design as an editing task within the latent space of a pretrained graph VAE. By introducing a novel compositional guidance mechanism that independently scales phenotypic targets and structural anchoring, PhAME resolves the problem of conflicting guidance signals. Consequently, it achieves SOTA performance across docking-score optimization, transcriptomic-conditioned generation, and Cell-Painting-driven Mode of Action generation while maintaining chemical validity and novelty. In future work, we plan to generalize this approach toward library design and hit generation tasks through multi-stage curriculum learning and to develop more robust biological oracles for evaluating de novo generated molecules.  
\section{Acknowledgments}
The work of Ł.J. and T.D. was carried out within the "AI-Based Virtual Screening Assistant" project (LIDER15/0033/2024) funded by the National Center for Research and Development (Poland) under the LIDER XV program. The work of S.M., D.R, and B.Z. was funded by "Interpretable and Interactive Multimodal Retrieval in Drug Discovery" project. The „Interpretable and Interactive Multimodal Retrieval in Drug Discovery” project (FENG.02.02-IP.05-0040/23) is carried out within the First Team programme of the Foundation for Polish Science co-financed by the European Union under the European Funds for Smart Economy 2021-2027 (FENG). We gratefully acknowledge Polish high-performance computing infrastructure PLGrid (HPC Center: ACK Cyfronet AGH) for providing computer facilities and support within computational grant no. PLG/2026/019504. Some experiments were performed on servers purchased with funds from the flagship project entitled “Artificial Intelligence Computing Core Facility” from the DigiWorld Priority Research Area within the Excellence Initiative – Research University program at Jagiellonian University in Krakow.

\bibliographystyle{abbrv}
\bibliography{references}

\appendix
\FloatBarrier
\section{Training Details}
 \paragraph{Python environment.} All experiments were carried out using Python 3.11.5. The key libraries are listed in Table~\ref{tab:libraries}. Other requirements are included in the code repository.

\begin{table}[h]
\centering
\caption{\textbf{Key libraries used in the experiments.}}
\label{tab:libraries}
\begin{tabular}{ll}
\toprule
\textbf{Library}                 & \textbf{Version} \\
\midrule
{PyTorch}                          & 2.10.0     \\ 
{PyTorch Lightning}                & 2.5.1            \\ 
{TorchGeometric}                   & 2.7.0           \\ 
{RDKit}                            & 2023.09.6        \\ 
{InfoAlign}                        & 0.1.1           \\ 
{Scikit-learn}                     & 1.8.0            \\ 
{Pandas}                           & 3.0.2            \\ 
{Numpy}                            & 1.26.4           \\ 
\bottomrule
\end{tabular}
\end{table}
 
\paragraph{Computational resources.}
 Experiments were conducted on a single NVIDIA Grace 72-core CPU @ 3.1 GHz and a single NVIDIA GH200 GPU with 96 GB of memory, using up to 32.0 GB of system RAM.

\label{hypers}
\FloatBarrier
\section{Details of logP Optimization}
\label{sec:app_logP_optimization}
\begin{figure}[htbp]
    \centering
\includegraphics[width=0.5\textwidth]{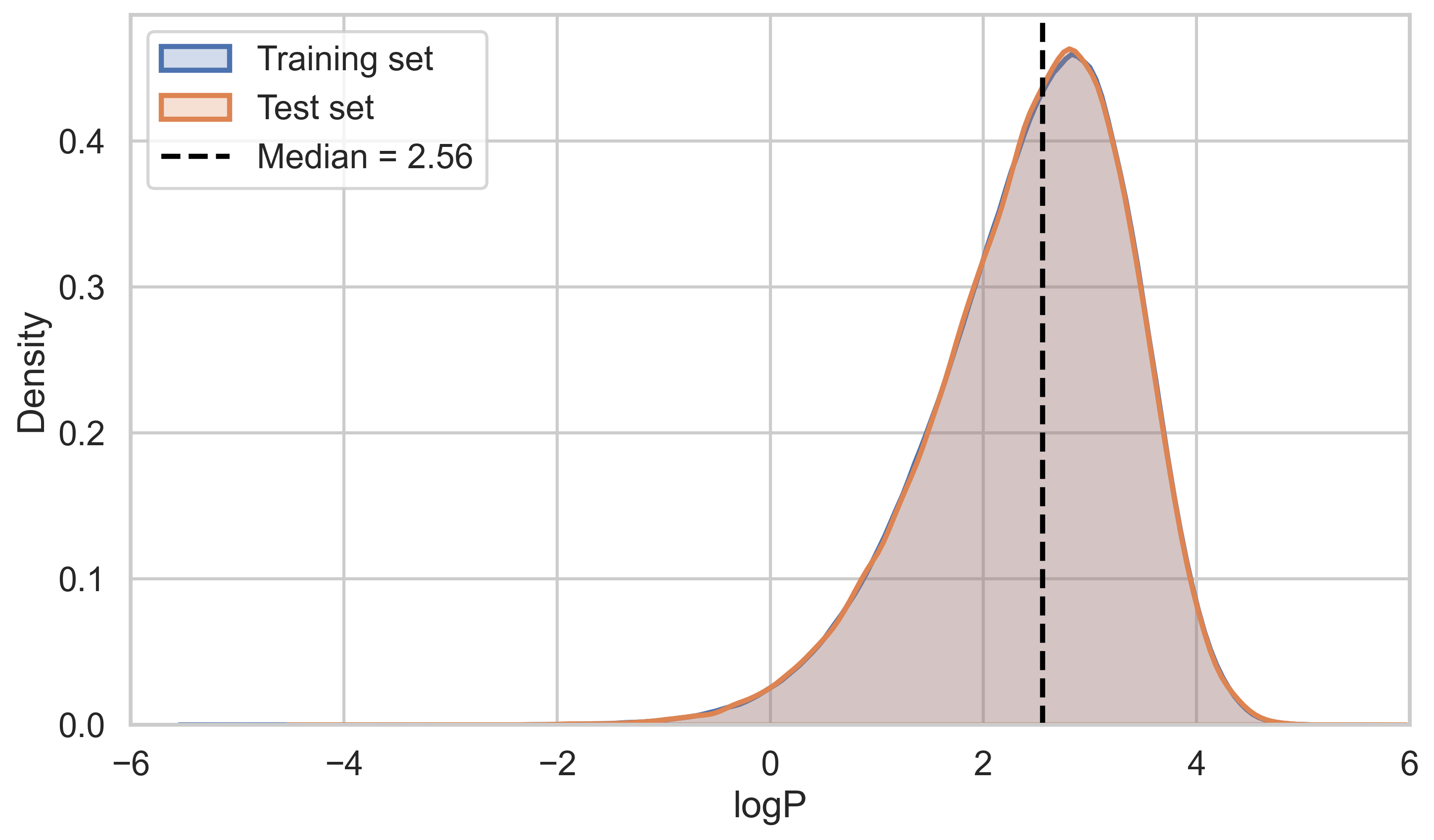}
    \caption{\textbf{Distribution of logP in the MOSES Dataset.} Distribution of logP values for the training and test sets derived from the MOSES dataset. The dashed vertical line denotes the global median logP of the train set.}
    \label{fig:log_distribution}
\end{figure}

\paragraph{Dataset.} 
To construct the training and testing datasets of pairs, we compute the logP values using the RDKit implementation of the Crippen model (i.e., \texttt{Crippen.MolLogP}). We then calculate the median logP value on the training set and use it as a threshold to split both sets. The training set contains an equal number of samples below and above this threshold, while the test set is split using the same value and undersampled to match the minority class. The distribution of logP values for both splits is shown in Figure~\ref{fig:log_distribution}.

\paragraph{Model.} 
For this task, we used the c-CFG (Compositional Classifier-Free Guidance) variant of PhAME. Unless otherwise stated, all reported results use $w_c = 6$ and $w_a = 3$. The model was trained for 1000 epochs with a learning rate of $1 \times 10^{-4}$, a batch size of 512, and 1000 diffusion steps. The alignment loss weight was set to $\gamma = 1$.

During inference, we initialize the process from a partially noised seed molecule at timestep $T = 900$, and perform the reverse diffusion process over the remaining steps.

\paragraph{Baselines implementation.} 
For our baselines, we adopted Mol-CycleGAN, which has a process most similar to our approach, it utilizes paired training. For DiGress, we used classifier guidance to steer the model toward the appropriate logP. However, it is worth noting that this model lacks an explicit mechanism for direct editing. PURE was trained following the setup described in its original paper. Crucially, our experimental setup does not allow for oracle-based optimization or filtering, as we rely solely on the provided labeled dataset. Therefore, to ensure a fair and equitable comparison, all post-processing steps were strictly disabled across all evaluated models.

\paragraph{Extended results.}
Table~\ref{tab:app_phame_ablation} summarizes the contribution of individual components. Standard Classifier-Free Guidance (denoted as CFG) achieves a reasonable trade-off, with moderate target success and similarity, resulting in limited performance in terms of \emph{NTS}. Employing Compositional Classifier-Free Guidance (denoted as c-CFG) with only the property signal active (\mbox{c-CFG}$(w_c{=}6, w_a{=}0)$) maximizes target success but severely degrades similarity, indicating that optimizing logP alone leads to substantial structural drift. Conversely, using only alignment guidance (\mbox{c-CFG}$(w_c{=}0, w_a{=}3)$) preserves high similarity but substantially reduces target success. Combining both signals (\mbox{c-CFG}$(w_c{=}6, w_a{=}3)$) yields a better balance between objectives. Removing the alignment loss decreases similarity and overall \emph{NTS}, indicating that it helps the model learn stronger structural alignment with the seed molecule. Finally, we demonstrate that the model can be aggressively steered to reach specific performance peaks. (\mbox{c-CFG}$(w_c{=}25, w_a{=}0)$) forces the model to achieve perfect target success. Applying a strong joint conditioning strategy (\mbox{c-CFG}$(w_c{=}12, w_a{=}3)$) yields the highest absolute \emph{NTS}, though the increased guidance signal comes at the cost of molecular validity.

As shown in Tables~\ref{tab:app_logp_results},~\ref{tab:app_phame_ablation}, quality metrics such as \emph{Val}, \emph{Uniq}, \emph{QED}, and \emph{SA} remain broadly comparable across methods, with no approach consistently dominating all criteria. One notable exception is standard CFG, which significantly reduces validity, indicating that stronger guidance can lead to chemically invalid structures. In contrast, methods incorporating alignment, including PhAME, maintain more stable validity while preserving competitive performance across the remaining quality metrics.
\begin{table*}
\centering
\setlength{\tabcolsep}{3.3pt}
\captionof{table}{\textbf{Comparison of methods for logP optimization with similarity preservation to the seed molecule.} Notably, PhAME achieves the highest score on the aggregated \emph{NTS} metric while maintaining strong, balanced performance across individual ones. Cells show mean $\pm$ std over 5 runs. }
\footnotesize
\setlength{\tabcolsep}{1.0pt}
\renewcommand{\arraystretch}{1.1}
\begin{tabular}{l ccccc|cccc}
\toprule
Method 
& Val $\uparrow$ & Uniq $\uparrow$ & QED $\uparrow$ & SA $\downarrow$ & Dir $\uparrow$
& Nov $\uparrow$ & Tgt $\uparrow$ & Sim $\uparrow$ & NTS $\uparrow$ \\
\midrule

Random target class  & 1.00\std{0.00} 
& 0.67\std{0.00} 
& 0.81\std{0.00} 
& 2.45\std{0.00} 
& 1.00\std{0.00} 
& 0.00\std{0.00} 
& 1.00\std{0.00} 
& 0.14\std{0.00} 
& 0.00\std{0.00} \\

Nearest target class
& 1.00\std{0.00} 
& 0.34\std{0.00} 
& 0.82\std{0.00} 
& 2.28\std{0.00} 
& 1.00\std{0.00} 
& 0.00\std{0.00} 
& 1.00\std{0.00} 
& 0.60\std{0.00} 
& 0.00\std{0.00} \\

VAE reconstruction 
& 0.91\std{0.00} 
& 0.53\std{0.00} 
& 0.80\std{0.00} 
& 2.70\std{0.00} 
& 0.53\std{0.00} 
& 0.99\std{0.00} 
& 0.19\std{0.00} 
& 0.59\std{0.00} 
& 0.11\std{0.00} \\

\midrule

Mol-CycleGAN 
& 0.72\std{0.00} 
& 0.90\std{0.01} 
& \textbf{0.82\std{0.01} }
&\textbf{2.57\std{0.07} }
& 0.71\std{0.03} 
& 0.99\std{0.00} 
& 0.51\std{0.04} 
& 0.15\std{0.00} 
& 0.08\std{0.01} \\

DiGrees & \underline{0.83\std{0.03}} 
& \textbf{0.99\std{0.00} }
& \underline{0.77\std{0.01} }
& 2.87\std{0.07} 
& \textbf{1.00\std{0.00}}
& \textbf{1.00\std{0.00}} 
& \textbf{1.00\std{0.00}}
& 0.13\std{0.00} 
& 0.13\std{0.00} \\

PURE & \textbf{0.86\std{0.00}} 
& \underline{0.96\std{0.00} }
& 0.54\std{0.01} 
& \underline{2.83\std{0.02} }
& 0.61\std{0.00} 
& \textbf{1.00\std{0.00}}
& 0.45\std{0.00} 
& \textbf{0.58\std{0.00} }
& 0.26\std{0.00} \\


PhAME  
& 0.71\std{0.03} 
& 0.88\std{0.01} 
& 0.74\std{0.01} 
& 2.96\std{0.05} 
& \underline{0.89\std{0.02} }
& \underline{0.99\std{0.00}} 
& \underline{0.78\std{0.03}} 
& \underline{0.39\std{0.02}} 
& \textbf{0.30\std{0.01}} \\

\bottomrule
\end{tabular}
\label{tab:app_logp_results}
\end{table*}
\begin{table*}
\centering
\captionof{table}{\textbf{Ablation of PhAME for logP optimization with similarity preservation 
to the seed molecule.} CFG refers to standard Classifier-Free Guidance, while c-CFG denotes compositional CFG. Cells show mean $\pm$ std over 5 runs. The results illustrate a clear trade-off across all metrics depending on the guidance scales.}
\footnotesize
\setlength{\tabcolsep}{1.0pt}
\renewcommand{\arraystretch}{1.1}
\begin{tabular}{l ccccc|cccc}
\toprule
Method 
& Val $\uparrow$ & Uniq $\uparrow$ & QED $\uparrow$ & SA $\downarrow$ & Dir $\uparrow$
& Nov $\uparrow$ & Tgt $\uparrow$ & Sim $\uparrow$ & NTS $\uparrow$ \\
\midrule

CFG(3)
& 0.42\std{0.02} 
& 0.89\std{0.01} 
& 0.70\std{0.01} 
& \textbf{2.80\std{0.06}}
& 0.81\std{0.01} 
& \underline{0.99\std{0.00}} 
& 0.68\std{0.01} 
& 0.40\std{0.00} 
& 0.27\std{0.01} \\

c-CFG(6, 0) 
& \textbf{0.87\std{0.02}}
& \textbf{0.97\std{0.01}}
& \underline{0.74\std{0.01}} 
& 2.95\std{0.07} 
& \underline{0.99\std{0.01}} 
& \textbf{1.00\std{0.00}}
& \underline{0.96\std{0.01}} 
& 0.13\std{0.00} 
& 0.12\std{0.00} \\

c-CFG(0, 3) 
& 0.73\std{0.03} 
& 0.88\std{0.00} 
& \textbf{0.76\std{0.01}} 
& 2.93\std{0.03} 
& 0.76\std{0.01} 
& \underline{0.99\std{0.00}} 
& 0.59\std{0.01} 
& \textbf{0.41\std{0.01}}
& 0.24\std{0.00} \\

c-CFG(6, 3) w/o $\mathcal{L}_{\text{align}}$ 
& \underline{0.80\std{0.02}} 
& \underline{0.94\std{0.00}} 
& \underline{0.74\std{0.00}} 
& \underline{2.81\std{0.02}} 
& 0.91\std{0.00} 
& \underline{0.99\std{0.00}} 
& 0.81\std{0.01} 
& 0.35\std{0.00} 
& 0.28\std{0.00} \\

c-CFG(6, 3) 
& 0.71\std{0.03} 
& 0.88\std{0.01} 
& \underline{0.74\std{0.01}} 
& 2.96\std{0.05} 
& 0.89\std{0.02} 
& \underline{0.99\std{0.00}} 
& 0.78\std{0.03} 
& \underline{0.39\std{0.02}} 
& \underline{0.30\std{0.01}} \\

c-CFG (25, 0)
& 0.67\std{0.10} 
& 0.86\std{0.03} 
& 0.62\std{0.07} 
& 2.90\std{0.15} 
& \textbf{1.00\std{0.00}} 
& \textbf{1.00\std{0.00}} 
& \textbf{1.00\std{0.01}} 
& 0.12\std{0.01} 
& 0.12\std{0.01} \\

c-CFG (12, 3)
& 0.67\std{0.04} 
& 0.89\std{0.01} 
& 0.71\std{0.02} 
& 2.99\std{0.07} 
& 0.93\std{0.02} 
& \underline{0.99\std{0.00}}
& 0.87\std{0.04} 
& 0.36\std{0.02} 
& \textbf{0.31\std{0.01}} \\

\bottomrule
\end{tabular}
\label{tab:app_phame_ablation}
\end{table*}

\FloatBarrier
\section{Details of Docking Score Optimization}
\label{sec:app_docking_score_optimization}

\paragraph{Three-stage curriculum.}
PhAME is trained in three stages on ZINC250k, following the pairing protocol of Section~\ref{sec:dataset}: at each stage, training pairs consist of molecules with QuickVina2 docking scores above and below a percentile-based threshold (non-binders vs.\ binders), with the most structurally similar pair selected across the threshold. Across stages, the threshold tightens from the 50th to the 90th and finally to the 99.8th percentile, so each stage fine-tunes the denoising network on a progressively higher-affinity property partition, a difficulty schedule rather than a self-distillation loop. Per-target percentile thresholds for ZINC250k are listed in Table~\ref{tab:percentiles_zinc}; per-stage training hyperparameters and target docking value (between the 99.8th and 100th percentile) are summarized in Table~\ref{tab:training_params}.

\paragraph{Model.} 
This is the only task that uses single-guidance CFG. All hyperparameters are listed in Table~\ref{tab:training_params}.

\paragraph{Hit thresholds.} As in MOOD~\citep{lee2023exploring}, the docking-score thresholds used in the \emph{novel hit ratio} are 10.0 for PARP1, 8.5 for FA7, 8.7845 for 5HT1B, 9.1 for JAK2, and 10.3 for BRAF.

\paragraph{Random VAE baseline.} \textit{Random VAE} samples 3{,}000 latents from $\mathcal{N}(0, I)$ in the DGVAE latent space and decodes them without guidance, providing a no-conditioning lower-bound on docking performance.

\begin{table}
\caption{\textbf{Percentile thresholds and numbers of compounds exceeding the threshold
for different targets in the ZINC250k dataset.}}
\label{tab:percentiles_zinc}
\makebox[\textwidth][c]{%
\begin{tabular}{cc|ccccc}
\toprule
Percentile & Count
& PARP1 & FA7 & 5HT1B & BRAF & JAK2 \\
 & ($\geq$ Thr.)
& Thr. & Thr. & Thr. & Thr. & Thr. \\
\midrule
0.10  & 227,000 & 6.4 & 5.5 & 5.9 & 6.6 & 6.4 \\
0.20  & 200,000 & 7.1 & 6.0 & 6.5 & 7.2 & 7.0 \\
0.50  & 126,000 & 8.2 & 6.7 & 7.9 & 8.2 & 7.9 \\
0.80  & 55,000  & 9.2 & 7.4 & 9.1 & 9.0 & 8.6 \\
0.90  & 27,000  & 9.7 & 7.8 & 9.6 & 9.5 & 9.0 \\
0.95  & 14,000  & 10.1& 8.1 & 10.0& 9.8 & 9.4 \\
0.99  & 3,000   & 10.9& 8.7 & 10.8& 10.5& 10.0 \\
0.998 & 600     & 11.5& 9.2 & 11.4& 11.1& 10.5 \\
1.0   & 1       & 14.9& 16.0 & 16.2& 14.7& 13.4 \\     
\bottomrule
\end{tabular}}
\end{table}
\begin{table}
\caption{\textbf{Training hyperparameters for different targets.}}
\label{tab:training_params}
\makebox[\textwidth][c]{%
\begin{tabular}{l|ccccc}
\toprule
 & PARP1 & FA7 & 5HT1B & BRAF & JAK2 \\
\midrule
Stage 1 threshold& 8.2& 7.4  & 7.9& 9.0   & 7.9\\
Stage 2 threshold& 10.9 & 8.5  & 10.8   & 10.5 & 10.0  \\
Stage 3 threshold& 11.5& 8.7  & 11.4& 11.0  & 10.5\\
\midrule
Stage 1 LR           & $1\mathrm{e}{-3}$ & $1\mathrm{e}{-3}$ & $1\mathrm{e}{-3}$ & $1\mathrm{e}{-3}$ & $1\mathrm{e}{-3}$ \\
Stage 1 epochs       & 1000 & 1000 & 1000 & 1000 & 1000 \\
Stage 2 LR           & $1\mathrm{e}{-3}$ & $1\mathrm{e}{-3}$ & $1\mathrm{e}{-3}$& $1\mathrm{e}{-4}$ & $1\mathrm{e}{-3}$ \\
Stage 2 epochs       & 500  & 500  & 1500 & 500  & 900\\
Stage 3 LR           & $1\mathrm{e}{-6}$ & $1\mathrm{e}{-4}$ & $1\mathrm{e}{-6}$& $1\mathrm{e}{-5}$ & $1\mathrm{e}{-6}$\\
Stage 3 epochs       & 500  & 800  & 500  & 800  & 500\\
\midrule
$\gamma\ $ & 0.01 & 0.01 & 0.01 & 0.01 & 0.01 \\
Batch size           & 512  & 512   & 512  & 512   & 512   \\
Cond. scale           & 2.5  & 2.0  & 3.5 & 2.5  & 3.0\\
Target            & 12.0 & 13.0  & 13.0  & 14.7  & 12.5  \\
\bottomrule
\end{tabular}}

\end{table}

\begin{figure}[htbp]
    \centering
    \includegraphics[width=1.0\textwidth]{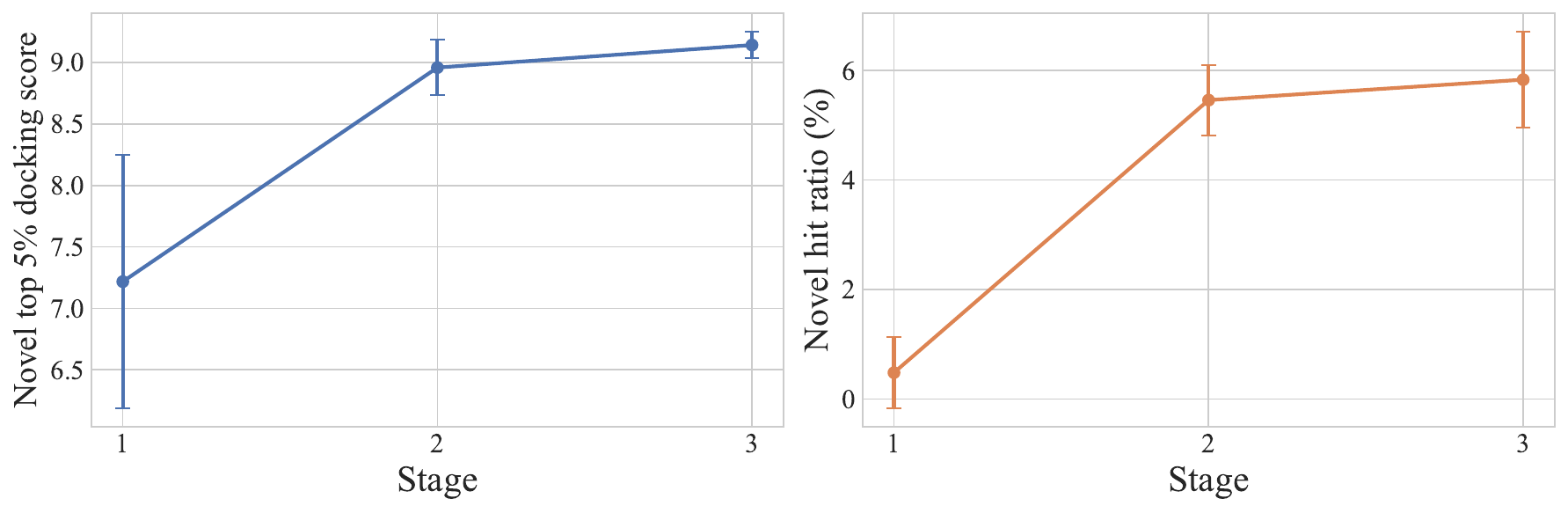}
    \caption{\textbf{Generative Performance Across Training Stages.} Novel top 5\% docking score and novel hit ratio (\%) across training stages. Values denote mean $\pm$ standard deviation over 5 runs.}
    \label{fig:ablation_docking_hit_ratio}
\end{figure}

\paragraph{Three-stage curriculum ablation.}
To attribute the gain reported in Section~\ref{sec:docking} to the iterative threshold-refresh curriculum rather than to the DGVAE backbone or compositional guidance alone, we evaluate a single-stage variant of PhAME on FA7: training only with the stage-1 threshold (50th percentile) and skipping the bootstrapped stage-2 and stage-3 fine-tuning. As can be seen in Figure~\ref{fig:ablation_docking_hit_ratio}, the single-stage variant achieves a novel hit ratio of $0.05$\% (vs.\ $5.75$\% for three-stage PhAME), confirming that the curriculum is the primary driver of the improvement.

\paragraph{Quantitative top-5 hits.}
\begin{table}
\caption{\textbf{Properties of the top 5 molecules ranked by docking score across all targets, satisfying the constraints $\mathbf{QED > 0.5}$, $\mathbf{SA < 5}$, and a maximum Tanimoto similarity to the training set of $\mathbf{< 0.4}$}}.
\label{tab:docking_scores_hits}
\centering
\begin{tabular}{c c c c c c}
\hline
 & ID & Similarity & Docking Score $\uparrow$ & QED $\uparrow$ & SA $\downarrow$ \\
\hline
\multirow{5}{*}{\rotatebox{90}{PARP1}} 
& 12  & 0.39 & 12.60 & 0.50 & 3.09 \\
& 105 & 0.30 & 12.50 & 0.52 & 4.05 \\
& 43  & 0.38 & 12.40 & 0.58 & 3.08 \\
& 107 & 0.38 & 12.40 & 0.57 & 3.63 \\
& 30  & 0.36 & 12.40 & 0.67 & 3.44 \\
\hline
\multirow{5}{*}{\rotatebox{90}{FA7}} 
& 25 & 0.38 & 10.60 & 0.67 & 3.75 \\
& 43 & 0.27 & 10.40 & 0.51 & 3.33 \\
& 64 & 0.33 & 10.20 & 0.51 & 4.02 \\
& 70 & 0.37 & 10.20 & 0.51 & 3.30 \\
& 91 & 0.38 & 10.10 & 0.68 & 3.23 \\
\hline
\multirow{5}{*}{\rotatebox{90}{5HT1B}} 
& 14  & 0.35 & 13.30 & 0.56 & 3.12 \\
& 29  & 0.38 & 12.30 & 0.60 & 3.45 \\
& 39  & 0.33 & 12.20 & 0.52 & 2.98 \\
& 142 & 0.37 & 12.20 & 0.63 & 3.76 \\
& 83  & 0.36 & 12.20 & 0.58 & 3.21 \\
\hline
\multirow{5}{*}{\rotatebox{90}{BRAF}} 
& 9   & 0.29 & 12.90 & 0.57 & 3.46 \\
& 17  & 0.36 & 12.80 & 0.52 & 3.53 \\
& 107 & 0.34 & 12.40 & 0.78 & 4.12 \\
& 128 & 0.39 & 12.30 & 0.67 & 3.42 \\
& 99  & 0.32 & 12.30 & 0.57 & 2.63 \\
\hline
\multirow{5}{*}{\rotatebox{90}{JAK2}} 
& 13  & 0.36 & 12.00 & 0.53 & 4.25 \\
& 46  & 0.31 & 11.70 & 0.56 & 3.03 \\
& 65  & 0.39 & 11.50 & 0.51 & 2.48 \\
& 93  & 0.38 & 11.40 & 0.62 & 3.93 \\
& 116 & 0.36 & 11.30 & 0.53 & 2.83 \\
\hline
\end{tabular}

\end{table}
Table~\ref{tab:docking_scores_hits} shows the quality metrics for the top five molecules for each hit. We report the QuickVina2 docking score, QED, SA, and the maximum Tanimoto similarity to any molecule in the ZINC250k training set.

\FloatBarrier
\section{Details of Transcriptomic Guided Generation}
\label{sec:app_transcriptomic_guided_generation}

\paragraph{Model.}
For this task, we used the c-CFG implementation of PhAME with condition guidance $w_c = 1$ and molecular-alignment guidance $w_a = 1$. Training was performed on the \texttt{MCF7} dataset using five random seeds $\{0,1,2,3,4\}$. For each molecule, up to three matched training pairs were used, selected with a minimum expression cosine distance of $0.5$/. For testing  $10\%$ of the data was held out.

The model was optimized for $4000$ epochs with a learning rate of $1 \times 10^{-4}$, batch size $512$, $1000$ diffusion steps, and $\mathcal{L}_{\text{align}} = 0.01$. Gene-expression conditions and molecular-alignment embeddings were projected into the latent conditioning space before being concatenated with the noisy latent representation and the time embedding. Specifically, the gene-expression condition was mapped from the full $978$-dimensional input space to the latent dimension using a linear projection layer, while the molecular embedding was projected through a two-layer MLP with a SiLU activation. 

As a seed molecule, we selected one of the ten most similar molecules based on cosine similarity between its gene expression and the target expression from the training set. During benchmarking, $100$ molecules were generated for each target. Each of the ten seed molecules was used $10$ times.

\paragraph{Data leakage handling.}
While largely following the original setup of Li \& Yamanishi~\citep{Li_2024_gxvaes,Hui_2025_smilesgen}, we introduce a minor but important modification to prevent potential data leakage. Specifically, we identify a subset of molecules that appear both in the training set and among the reference ligands used for evaluation. These overlapping molecules are removed from the training data to ensure a strict separation between training and evaluation. This filtering reduces the training set size from 13,755 to 13,432 molecules (less than 3\% of the data).

\paragraph{Dataset creation.}
We construct a paired compound–transcriptome dataset as described in Section~\ref{sec:dataset}. Unlike the property-based setting, transcriptomic data does not admit a natural threshold for partitioning, and we therefore adopt a similarity-based pairing strategy. Specifically, we compute cosine similarity between normalized gene expression profiles. For each molecule, we select the most structurally similar candidates based on Tanimoto similarity over ECFP fingerprints, while enforcing transcriptomic dissimilarity, $\cos(g_i, g_j) < 0.5$, where $g$ denotes the gene expression profile. This results in training pairs that are structurally similar but induce distinct transcriptional responses.





\FloatBarrier
\section{Details of Cell Phenotype Guided Generation}
\label{sec:app_cell_phenotype_guided_generation}
\paragraph{Dataset.}

PhAME was trained via a two-stage strategy that uses distinct threshold parameters to construct both the general and the fine-tuning datasets. For the general dataset, the threshold was set to $\tau_{\text{cos}}= -0.3$. For fine-tuning, we selected a much stricter threshold: $\tau_{\text{cos}} = -0.6$. In both stages, we first isolated all candidates whose phenotypic cosine similarity fell below $\tau_{\text{cos}}$ (computed on features extracted from CellProfiler). Among these, we selected the pair with the highest Tanimoto similarity. This protocol is crucial because mining a sufficiently large dataset of highly divergent biological responses and strongly similar chemical structures is difficult. The proposed regime provides the volume needed for stable pre-training and then strictly fine-tunes the model on the desired dataset. As outlined, the dataset was created from the CP-JUMP database, with features extracted from CellProfiler. The test set was generated using the CP-JUMP database and intersected with ChEMBL2K and the Broad Drug Repurposing Hub. Crucially, to ensure a perfectly balanced and unbiased evaluation (without intersection of MoA classes), we restricted our focus to seven distinct MoA classes comprising exactly 8 representative molecules per class: \textit{acetylcholine receptor antagonist}, \textit{adrenergic receptor agonist}, \textit{adrenergic receptor antagonist}, \textit{cyclooxygenase inhibitor}, \textit{dopamine receptor antagonist}, \textit{histamine receptor antagonist}, and \textit{serotonin receptor antagonist}.

\paragraph{Feature provenance.}
At inference, PhAME is conditioned on the per-MoA centroid (mean over the 8 reference compounds in each class) of CellProfiler~\citep{carpenter2006cellprofiler} morphological features computed on each compound's CP-JUMP~\citep{chandrasekaran2023align} profiles. We use the specific preprocessed feature flavor distributed by the InfoAlign authors~\citep{liu2025learning} on their public HuggingFace repository, which we denote \textit{CellProfiler} in our ablations. A head-to-head comparison of CellProfiler against two alternative cellular feature extractors (\textit{DeepProfiler} and \textit{uniDino}) is reported in the Morphology Representation ablation below.

\paragraph{Training details and baselines.}
For the training of the other baselines CPMolGAN and GFlowNet, we strictly follow their originally proposed training procedures and hyperparameter configurations. We omit the recent method MGMG~\citep{tang2025mgmg} from our evaluation framework, as its codebase is not publicly available. PhAME was optimized for $2000$ epochs with a learning rate of $1 \times 10^{-5}$ in fine-tune stage and $1 \times 10^{-3}$ in general stage, batch size $512$, $1000$ diffusion steps, with early stopping for $50$ epochs, warmup $200$ epochs, dropout $0.1$. However, because this specific evaluation focuses on \textit{de novo} generation rather than localized structural optimization, we adjusted the training and inference parameters accordingly. During training, we incorporated EMA (decay factor of 0.999) optimization to stabilize training and reduced the structural-alignment weight $\gamma$ to 0.00001. We set the structural guidance scale to $w_a=0$ to allow unconstrained exploration, and the conditioning scale to $w_c=34$ to enforce a strong cellular signal, a value determined empirically to yield the highest biological fidelity. 

Specifically, for each of the 7 MoA classes, the generative process was conditioned directly on the pre-computed biological centroid of that specific class. To ensure statistical robustness in our evaluation, we conducted sampling across 5 independent random seeds, generating 10 valid compound samples per representative molecule, yielding 80 generated molecules per MoA class in total.

\paragraph{Metrics.}

Let $\hat m \in \mathcal{M}$ denote a generated molecule with target MoA $C^* \in \mathcal{Y} = \{C_1, \ldots, C_7\}$, and let $\mathcal{R}_C \subset \mathcal{M}$ collect the reference compounds annotated with MoA $C$ ($|\mathcal{R}_C|{=}8$ in our balanced benchmark). All three metrics below share a representation-specific pairwise similarity $s(\cdot, \cdot) \in [0, 1]$ between two molecules: for the ECFP space, $s$ is Tanimoto similarity over Morgan fingerprints; for the learned spaces (CLOOME, InfoAlign), $s$ is cosine similarity over the corresponding embeddings of the two molecules.

\begin{itemize}
\item \textbf{Top-1 Cluster Accuracy} scores each class by measuring the mean similarity between the generated molecule and all individual members of reference class in evaluation representation,
\[
\sigma(\hat m, C) \;=\; \frac{1}{|\mathcal{R}_C|} \sum_{r \in \mathcal{R}_C} s\!\left(\hat m,\, r\right),
\]

and reports the fraction of generations for which the target attains the highest score, i.e.\ $C^{*} = \arg\max_{C \in \mathcal{Y}} \sigma(\hat m, C)$.

\item \textbf{MoA Retrieval Rate @$k$} relaxes Top-1 to admit the target anywhere in the top $k$ ranked classes, and replaces the centroid-based score above with a top-$n$ mean of pairwise similarities that focuses on each class's most similar references:
\[
\rho(\hat m, C) \;=\; \frac{1}{n}\sum_{r \in \mathrm{Top}_n(\hat m,\, \mathcal{R}_C)} s(\hat m, r), \qquad n = 3.
\]
Sorting classes by $\rho(\hat m, \cdot)$ in descending order yields $\mathrm{rank}(C^{*})$; the metric is the fraction of generations with $\mathrm{rank}(C^{*}) \leq k$.

\item \textbf{kNN Accuracy @$k$} discards the per-class structure and runs a $k$-nearest-neighbour classifier over the pooled reference set $\mathcal{R} = \bigcup_{C \in \mathcal{Y}} \mathcal{R}_C$. It takes the $k$ references most similar to $\hat m$ under $s$ and assigns the majority MoA among them as the predicted label; the metric is the fraction of generations whose prediction equals $C^{*}$.
\end{itemize}

\paragraph{Extended Results} 
\begin{table}[t]
\centering
\caption{\textbf{MoA classification of de novo generated molecules across three representation spaces.} Values are percentages (mean${}_{\pm\text{std}}$ over 5 seeds). Best in bold, second-best underlined. PhAME achieves the best performance in \textbf{20 out of the 21} evaluated combinations of metrics and representation spaces.}
\label{tab:app_cell}
\scriptsize
\renewcommand{\std}[1]{{\tiny\,$\pm$#1}}
\setlength{\tabcolsep}{6pt}
\renewcommand{\arraystretch}{1.1}

\resizebox{\textwidth}{!}{%
\begin{tabular}{ll c ccc ccc}
\toprule
& & \multicolumn{1}{c}{\textbf{Top-1}}
& \multicolumn{3}{c}{\textbf{MoA Retrieval Rate @$k$}~$\uparrow$}
& \multicolumn{3}{c}{\textbf{kNN Accuracy @$k$}~$\uparrow$} \\
\cmidrule(lr){3-3} \cmidrule(lr){4-6} \cmidrule(lr){7-9}
& \textbf{Model}
& \textbf{Cluster Acc}~$\uparrow$
& \textbf{@1} & \textbf{@2} & \textbf{@3}
& \textbf{@1} & \textbf{@3} & \textbf{@5} \\
\midrule
\multirow{4}{*}{\rotatebox[origin=c]{90}{CLOOME}}
 & DGVAE        & 14.6 \std{1.2} & 13.7 \std{1.8} & 29.1 \std{2.0} & 42.9 \std{2.5} & 13.6 \std{2.0} & 14.0 \std{2.7} & 14.4 \std{1.5} \\
 & CPMolGAN     & 15.4 \std{1.5} & 14.8 \std{1.4} & 28.8 \std{1.8} & 42.4 \std{2.2} & 14.9 \std{1.3} & 15.2 \std{1.3} & 15.2 \std{1.8} \\
 & GFlowNet     & \textbf{21.2 \std{1.2}} & \textbf{19.4 \std{2.5}} & \underline{34.4 \std{2.1}} & \underline{51.5 \std{2.2}} & \underline{18.0 \std{1.1}} & \underline{18.6 \std{1.1}} & \underline{19.3 \std{1.2}} \\
 & PhAME & \underline{17.8 \std{1.1}} & \textbf{19.4 \std{0.5}} & \textbf{38.2 \std{0.9}} & \textbf{53.6 \std{2.1}} & \textbf{20.5 \std{1.2}} & \textbf{20.4 \std{1.0}} & \textbf{20.4 \std{0.7}} \\
\midrule
\multirow{4}{*}{\rotatebox[origin=c]{90}{InfoAlign}}
 & DGVAE        & 13.4 \std{0.5} & 14.4 \std{1.0} & 28.6 \std{1.9} & 43.2 \std{3.2} & 14.0 \std{1.7} & 14.3 \std{1.8} & 14.2 \std{1.3} \\
 & CPMolGAN     & 13.1 \std{0.9} & 13.5 \std{1.2} & 26.9 \std{0.7} & 39.6 \std{1.0} & 13.5 \std{1.2} & 13.1 \std{1.1} & 13.6 \std{0.8} \\
 & GFlowNet   & \underline{17.1 \std{0.7}}  & \underline{16.3 \std{1.3}} & \underline{32.4 \std{0.8}} & \underline{50.1 \std{1.7}} & \underline{14.6 \std{0.5}} & \underline{15.7 \std{0.6}} & \underline{15.9 \std{1.3}} \\
 & PhAME & \textbf{28.7 \std{1.7}} & \textbf{28.7 \std{2.7}} & \textbf{45.2 \std{1.8}} & \textbf{58.0 \std{1.0}} & \textbf{25.4 \std{2.2}} & \textbf{27.0 \std{1.8}} & \textbf{28.3 \std{1.4}} \\
\midrule
\multirow{4}{*}{\rotatebox[origin=c]{90}{ECFP}}
 & DGVAE        & 14.0 \std{0.9} & 14.2 \std{2.0} & 28.5 \std{1.1} & 42.7 \std{2.1} & \underline{14.5 \std{0.6}} & \underline{14.6 \std{1.0}} & 13.7 \std{1.6} \\
 & CPMolGAN     & 14.3 \std{1.0} & 13.4 \std{1.3} & 28.8 \std{2.4} & 42.0 \std{0.9} & 12.9 \std{1.2} & 13.5 \std{1.4} & 14.1 \std{1.4} \\
 & GFlowNet     & \underline{17.0 \std{1.8}} & \underline{15.1 \std{1.3}} & \underline{29.0 \std{1.6}} & \underline{46.1 \std{2.1}} & 12.7 \std{1.1} & 14.4 \std{1.3} & \underline{15.3 \std{1.6}} \\
 & PhAME & \textbf{27.2 \std{1.8}} & \textbf{23.0 \std{1.6}} & \textbf{40.4 \std{2.3}} & \textbf{56.2 \std{2.9}} & \textbf{20.6 \std{1.2}} & \textbf{21.2 \std{1.6}} & \textbf{23.0 \std{2.2}} \\
\bottomrule
\end{tabular}%
}
\end{table}

Table~\ref{tab:app_cell} demonstrates a contrast in performance across the three representation spaces. Both DGVAE and CPMolGAN achieve results near a random-guess baseline ($\sim$14.3\% Top-1 Cluster Accuracy) regardless of the embedding used. Although GFlowNet performs well within CLOOME, its performance drops when evaluated in InfoAlign and ECFP spaces. PhAME is the \emph{only} method that clears chance in all three spaces simultaneously, and the gap is largest exactly where the biological signal is richest: $+68\%$ relative over GFlowNet on InfoAlign Top-1 Cluster Accuracy ($0.287$ vs.\ $0.171$) and $+60\%$ on ECFP ($0.272$ vs.\ $0.170$), with similar gains reflected in MoA Retrieval Rate @$k$ and kNN Accuracy @$k$. On CLOOME, PhAME trails GFlowNet on Top-1 ($0.178$ vs.\ $0.212$); PhAME's lead reappears once we move to deeper metrics (MoA Retrieval Rate @1/2/3, kNN Accuracy @1/3/5). Because the InfoAlign and ECFP advantages hold across two structurally unrelated representations, the improvements reflect a true understanding of structure-activity relationships rather than merely adapting to a single representational topology. 

For completeness,  Table~\ref{tab:stats_cell} reports extended statistics on molecular quality and diversity metrics (QED, SA, Uniqueness and Novelty). These results should be contextualized with the primary Table~\ref{tab:app_cell}. While baselines like DGVAE and CPMolGAN achieved slightly better QED or SA, they do not perform effectively on the actual MoA retrieval task. Importantly, PhAME still outperforms its primary competitor, GFlowNet, in both QED and SA.

\begin{figure}
    \centering
    \includegraphics[width=\linewidth]{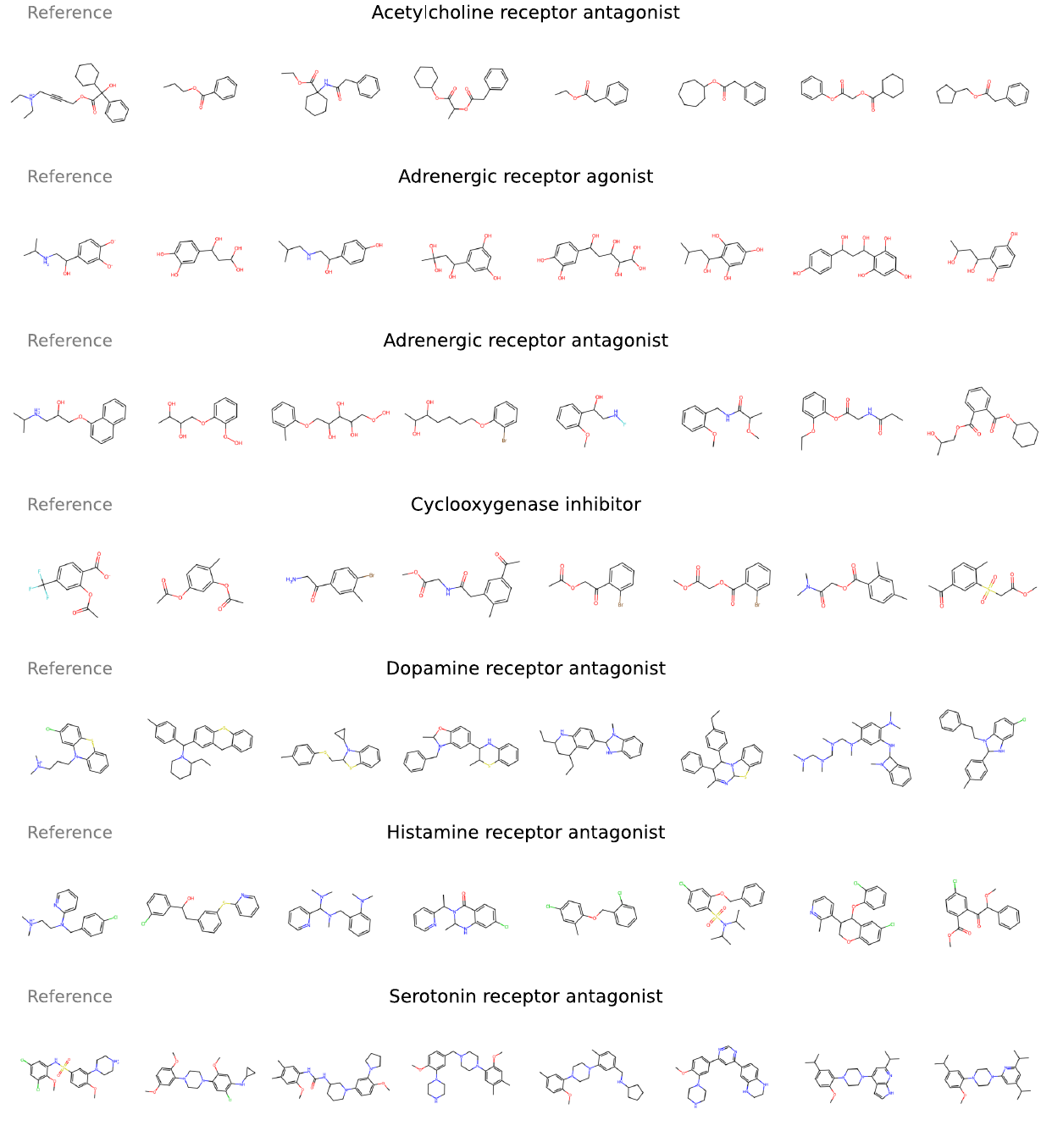}
    \caption{\textbf{Qualitative Assessment of PhAME-Generated molecules.} Qualitative examples of molecules generated by PhAME, conditioned by MoA embedding centroids. Each row presents a different Mode of Action. The first column captioned as Reference represents one of the selected molecules from our dataset with a known MoA, while the remaining columns visualize the nearest generated molecules based on Tanimoto similarity to any molecule in the MoA cluster. Each class contains specific, distinct molecular structures.}
    \label{fig:moa_}
\end{figure}

\paragraph{Qualitative MoA guided molecules}
Figure~\ref{fig:moa_} shows selected molecules conditioned by cell morphology described in Section~\ref{sec:phenotype}. To analyze the structural relevance, we visualize the nearest generated molecules based on Tanimoto similarity to any of the molecules from each MoA.
\begin{table}
\centering
\caption{\textbf{Evaluation of quality and diversity generated molecules for Cell Phenotype task.} Average metrics for cell-phenotype-guided generation, evaluated across five distinct seed molecules.}
\label{tab:stats_cell}
\begin{tabular}{lcccc}
\toprule
Method & QED $\uparrow$ & SA $\downarrow$ & Uniqueness $\uparrow$ & Novelty $\uparrow$ \\
\midrule
DGVAE  & \textbf{0.797 \std{0.001}} & \textbf{2.515 \std{0.014}} & \textbf{1.000 \std{0.000}} & \textbf{1.000 \std{0.001}} \\
CPMolGAN  & 0.607 \std{0.010} & \underline{3.042 \std{0.037}} & \textbf{1.000 \std{0.000}} & \textbf{1.000 \std{0.000}} \\
GFlowNet  & 0.649 \std{0.005} & 4.457 \std{0.024} & \underline{0.999 \std{0.002}} & \textbf{1.000 \std{0.000}} \\
PhAME & \underline{0.674 \std{0.003}} & 3.233 \std{0.027} & 0.960 \std{0.007} & \textbf{1.000 \std{0.001}} \\
\bottomrule
\end{tabular}
\end{table}

\section{Details of Cell Phenotype Optimization}
\label{app:cell_phenotype_optimization}
\paragraph{Setup and target selection.}
To evaluate the capacity of H2L optimization, we formulated a recovery task of drugs. We selected three different molecules with known MoAs.
The selected target therapeutics and their respective MoAs are as follows:
\begin{itemize}
    \item \textbf{Aspirin:} \textit{Cyclooxygenase inhibitor}
    \item \textbf{Clofibrate:} \textit{PPAR receptor agonist}
    \item \textbf{Dyphylline:} \textit{Adenosine receptor antagonist}
\end{itemize}
For each target therapeutic, we selected a set of structurally similar "hit" molecules to serve as starting seeds in order to evaluate the model's performance across varying levels of optimization difficulty. Crucially, to strictly prevent data leakage, the target molecule was entirely removed from the reference dataset prior to inference. Specifically, when constructing the conditioning signal, the target MoA centroid was computed exclusively from the phenotypic representations of the remaining class members, ensuring the model had no direct access to the target drug's biological signature.

\paragraph{Evaluation.}
PhAME was evaluated using the configuration from Appendix~\ref{sec:app_cell_phenotype_guided_generation}, with guidance scales specifically tuned for structural editing $w_a=1$ and $w_c=14$. Unlike \textit{de novo} generation, which starts from pure Gaussian noise, we start the process from a noised seed molecule at the corresponding timesteps $T=400$ and $T=500$. We found that each seed has a different optimal initial noise level: the level of noise depends on the structural similarity between lead and hit. For instance, among our visualized examples, only one seed molecule (the aspirin hit presented in row 2, column 2) required a significantly higher initial noise level ($T=700$) to successfully bridge the structural gap. For this evaluation, we generated a total of 100,000 candidate samples. On the same setup, we conducted tests with the GFlowNet model. For the seed molecules presented in Figure~\ref{fig:h2l_cell} conditioned with the proper morphology, we could not recover any of the drugs.

\begin{figure*}
    \centering
    \includegraphics[width=0.7\linewidth]{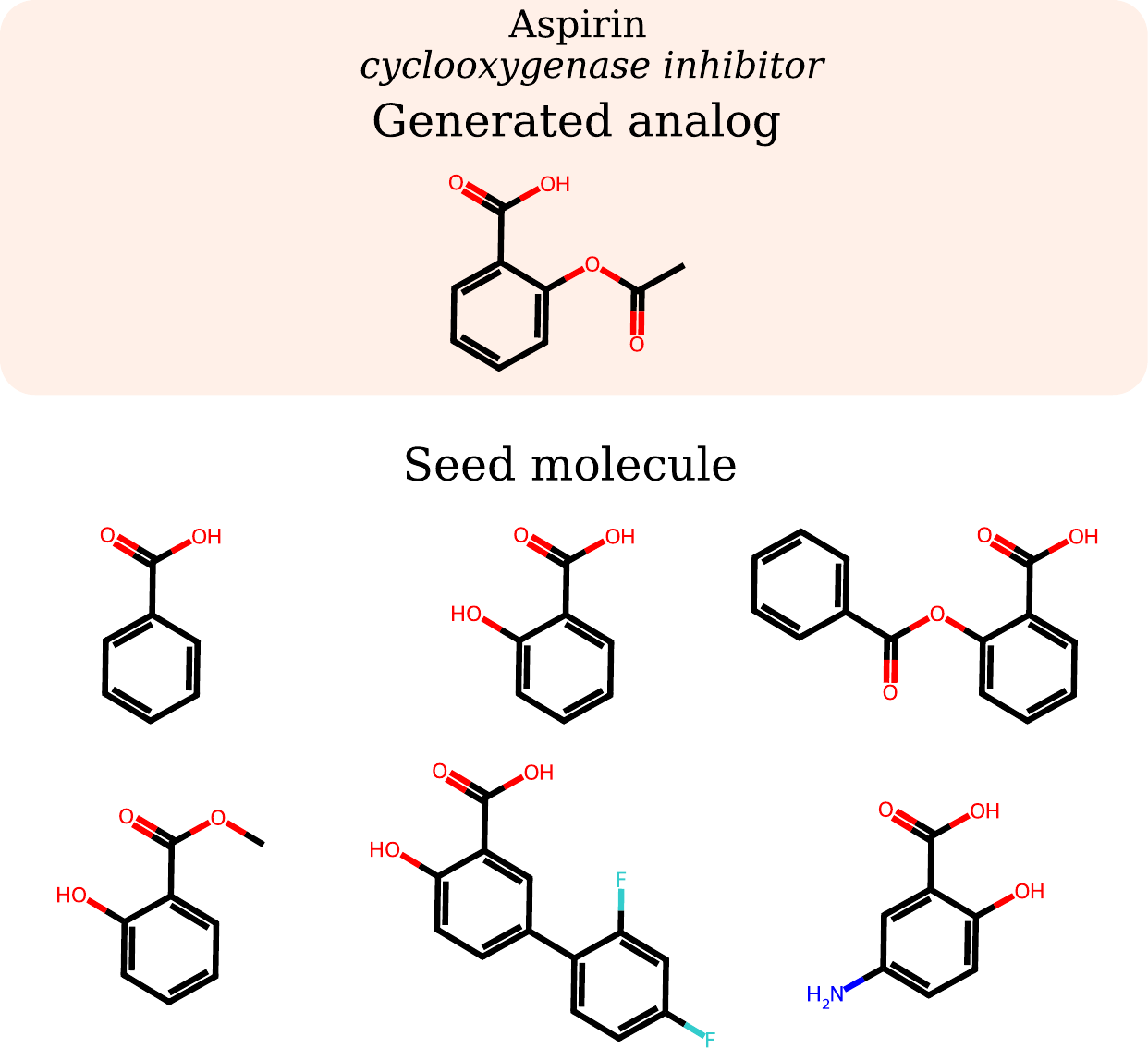}
    \caption{\textbf{Recovery of Aspirin.} Qualitative examples of PhAME recovering Aspirin Lead (Generated Analog). Six distinct Hits (Seed molecule) were evaluated using structural anchor and biological conditioning of cyclooxygenase inhibitor.}
    \label{fig:app_aspirin}
\end{figure*}

\begin{figure*}
    \centering
    \includegraphics[width=0.7\linewidth]{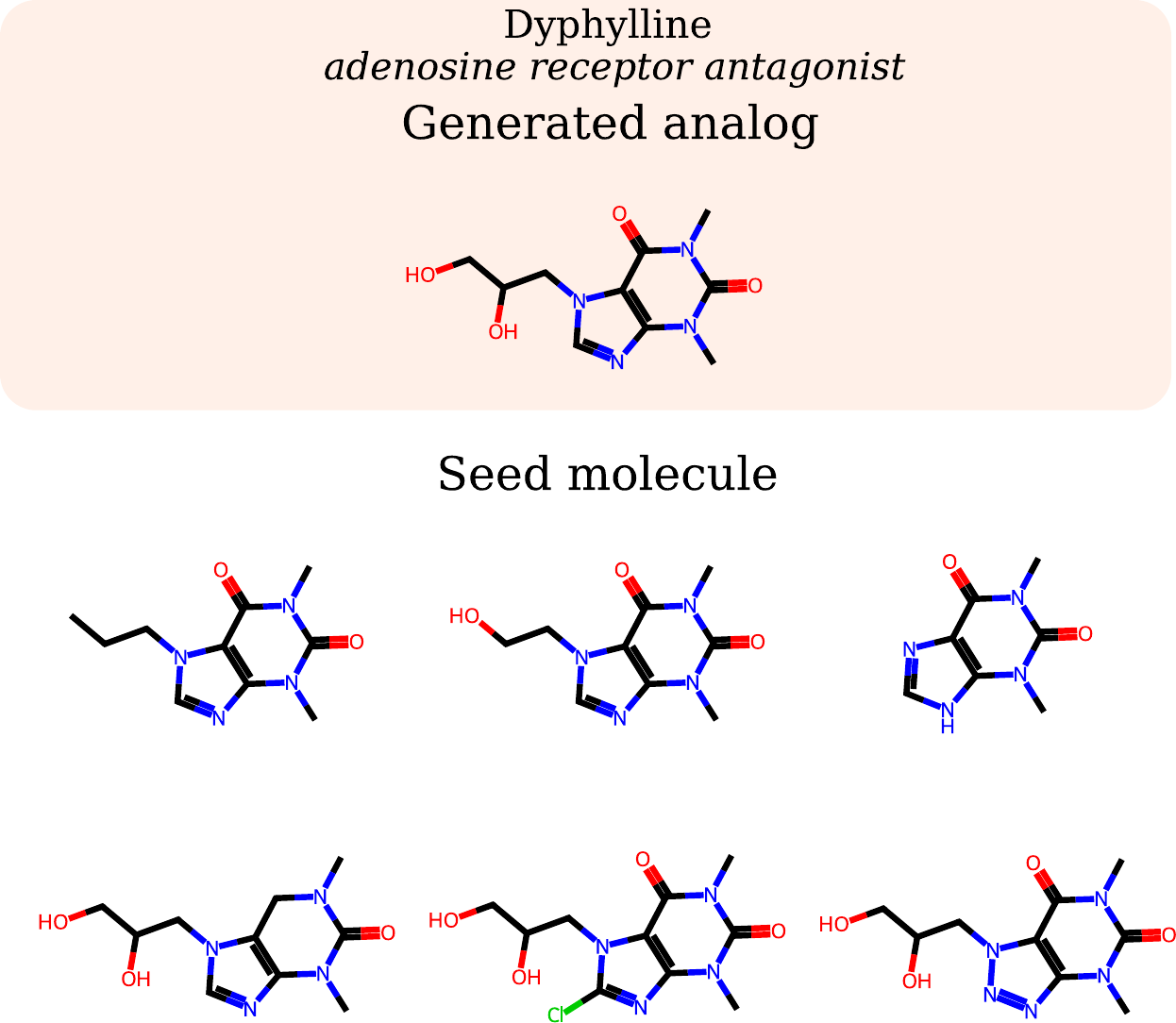}
    \caption{\textbf{Recovery of Dyphylline.} Qualitative examples of PhAME recovering Dyphylline Lead (Generated Analog). Six distinct Hits (Seed molecule) were evaluated using structural anchor and biological conditioning of adenosine receptor antagonist.}
    \label{fig:app_dyphylline}
\end{figure*}

\begin{figure*}
    \centering
    \includegraphics[width=0.7\linewidth]{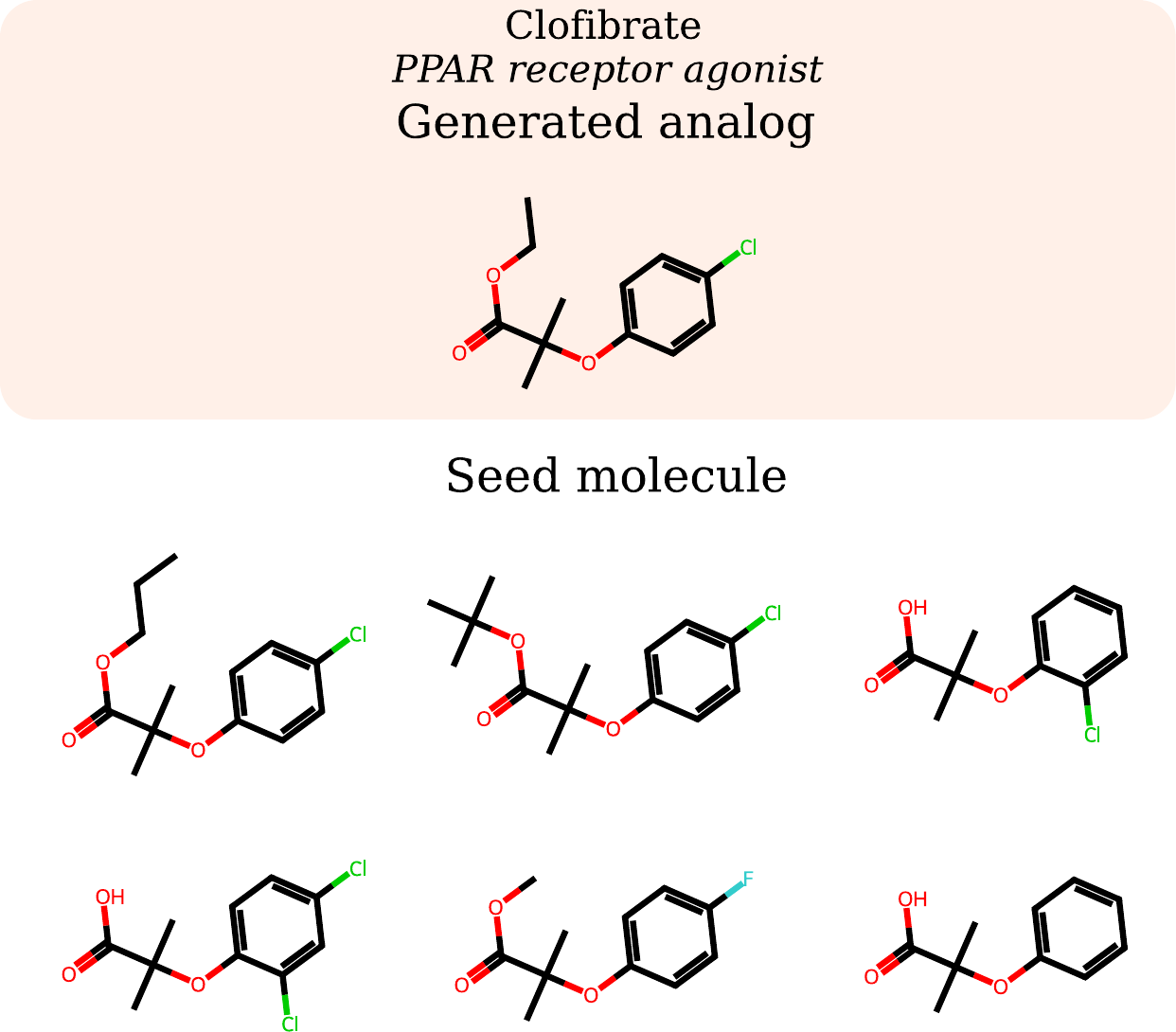}
    \caption{\textbf{Recovery of Clofibrate.} Qualitative examples of PhAME recovering Clofibrate Lead (Generated Analog). Six distinct Hits (Seed molecule) were evaluated using structural anchor and biological conditioning of PPAR receptor agonist.}
    \label{fig:app_clofibrate}
\end{figure*}

To illustrate PhAME's structural editing capabilities, we present extended qualitative visualizations in Figures~\ref{fig:app_aspirin},~\ref{fig:app_dyphylline},~\ref{fig:app_clofibrate}. We provide comprehensive trajectories for each of the three targeted therapeutics: aspirin, clofibrate, and dyphylline. For each target lead, we display successful recoveries starting from six distinct hit molecules. These examples have varying complexity in the required transformation, ranging from minimal, straightforward functional-group edits to more advanced ones.

\FloatBarrier
\section{Ablations}
\subsection{Morphology Representation}
To evaluate the robustness of our model to different conditioning representations from different biological manifolds, we conducted an ablation study comparing various cellular representations. Specifically, we selected three different types of feature representations extracted from the CP-JUMP dataset. Our primary cell morphology signal, denoted \textit{CellProfiler}, corresponds to the exact same CellProfiler features utilized by InfoAlign. We compared it against two alternative representations: \textit{DeepProfiler}~\citep{moshkov2022cpcnn} features extracted from a CNN-based model, retrieved from the official CP-JUMP repository and \textit{uniDino}~\citep{morelli2025unidino} representations extracted from the uniDino model vision transformer. Crucially, to ensure a fair comparison across these three modalities, we restricted this study to a single experimental source (\textit{Source 2} from CP-JUMP), which reduces the dataset size by approximately 50\% compared to our main experiments. 

\begin{figure}[htbp]
    \centering
    \includegraphics[width=1.0\textwidth]{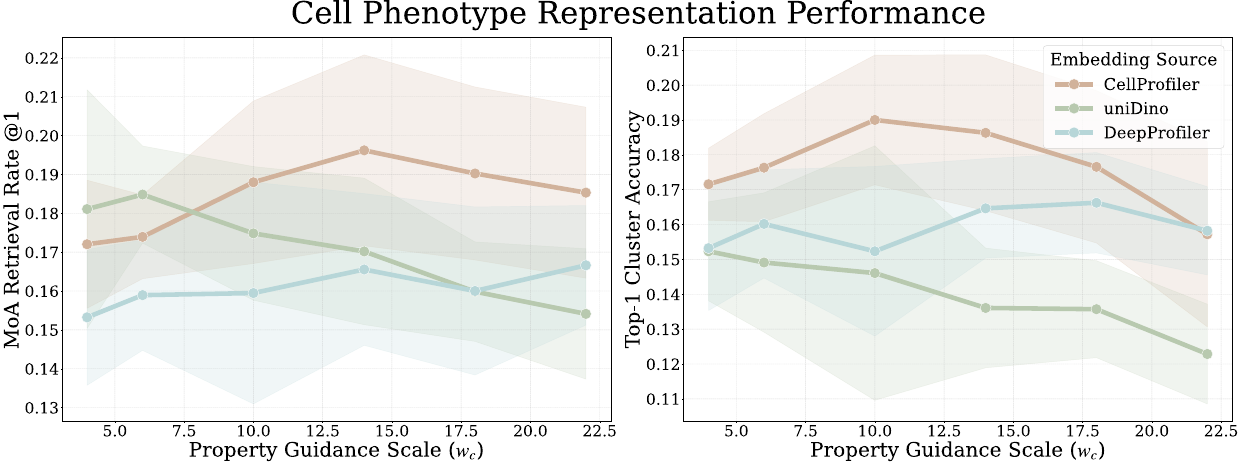}
    \caption{\textbf{Performance of Phenotypic Feature Extractors}. Comparison of cell phenotypic representations across different feature extractors: CellProfiler, UniDino and DeepProfiler. Performance is evaluated using the Top-1 Cluster Accuracy and MoA Retrieval Rate, metrics described in Appendix~\ref{sec:app_cell_phenotype_guided_generation}. The CellProfiler representation achieves the best results across its head-to-head competitors as a signal for conditioning ($w_c$).} 
    \label{fig:ablations_cell_representation}
\end{figure}

Figure~\ref{fig:ablations_cell_representation} highlights the advantage of our chosen representation. It shows that \textit{CellProfiler} achieves the highest overall performance on the Top-1 Cluster Accuracy and MoA Retrieval Rate@k metrics, averaged across all evaluation spaces (CLOOME, InfoAlign, and ECFP).

\subsection{Structural Alignment Representation}
\begin{figure}[htbp]
    \centering
    \includegraphics[width=1.0\textwidth]{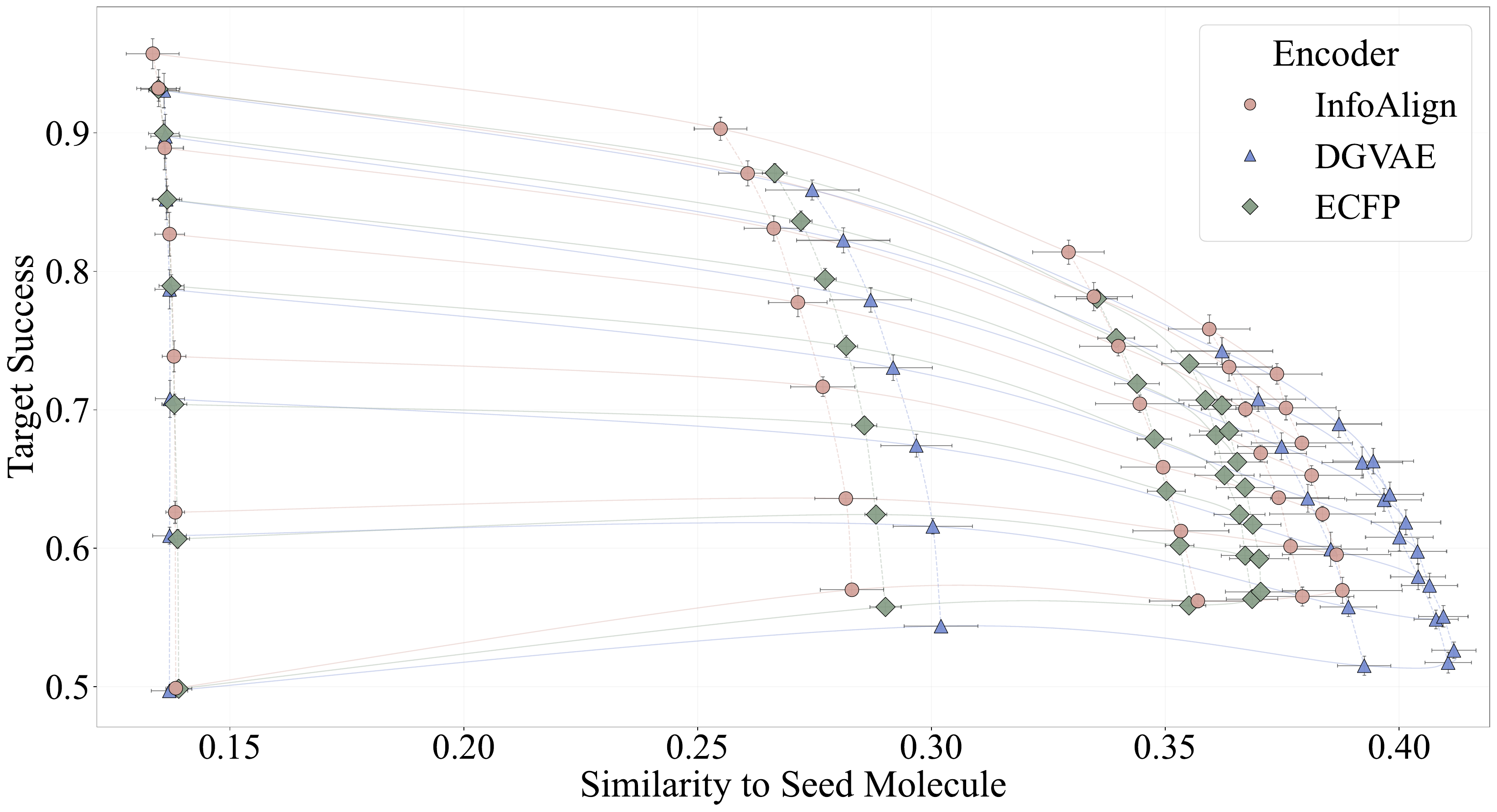}
    \caption{\textbf{Impact of Representations on Controllable Generation.} Trade-off between Target Success and Similarity to Seed Molecule under varying guidance strengths for different type encoded representation for conditioning $w_a$. Increasing property guidance ($w_c$) improves target success, while increasing seed guidance ($w_a$) enhances similarity to the input molecule. InfoAlign achieves the highest Target Success, showing only a marginal decrease in Similarity to Seed Molecule relative to DGVAE, which excels in seed similarity but struggles with Target Success. In contrast, ECFP shows weak performance across both criteria, further highlighting the advantage of InfoAlign. The layout and interpretation of this plot are identical to those of Figure\ref{fig:guidance_optimization}}
    \label{fig:ablations_pareto_encoders}
\end{figure}

To validate our choice of structural alignment representation $z_{\text{align}}$ in Equations~\ref{eq:L_align},~\ref{eq:multi-cfg}, we evaluated PhAME on three distinct molecular representation embedding spaces: Morgan Fingerprint (ECFP), DGVAE latent space and InfoAlign embeddings in the setup of logP optimization described in Appendix~\ref{sec:app_logP_optimization}, with limited dataset to 20\% of original size. As shown in Figure~\ref{fig:ablations_pareto_encoders}, InfoAlign slightly decreases similarity alignment to achieve superior property-target alignment. We accept a minor, negligible trade-off in structural similarity because InfoAlign offers a much richer representation. 



\section{Limitations and Broader Impact}
\label{sec:limitations_impact}

\paragraph{Limitations.}
While PhAME mitigates the trade-offs in standard classifier-free guidance, the framework remains sensitive to extreme guidance scales, where overly strong structural guidance can degrade the chemical validity of the generated molecules. Additionally, the model's performance relies on the availability and quality of high-dimensional paired data, such as Cell Painting morphologies and L1000 transcriptomic profiles.

\paragraph{Broader impact.}
PhAME represents an advancement in machine learning-driven drug design, directly supporting the pharmaceutical industry's shift from single-target assays toward system-level phenotypic drug discovery. By effectively guiding molecules toward complex biological signatures while anchoring them to known chemical seeds, this framework directly addresses the bottlenecks of Hit-to-Lead (H2L) and Lead Optimization (LO) pipelines. Accelerating this iterative refinement process has the potential to significantly reduce the computational and financial costs associated with reshaping screening hits into potent and safe drug candidates.


\newpage

\end{document}